\documentclass[twocolumn]{iosart2c}

\usepackage{dcolumn}
\usepackage{graphicx}
\usepackage{multirow}
\usepackage{subfigure}
\usepackage{lettrine}
\usepackage{algorithm}
\usepackage{algorithmicx}
\usepackage{algpseudocode}

\usepackage{color}
\usepackage{ulem}

\ifx00  
 \newcommand*{\add}[1]{{#1}}
 \newcommand*{\id}[1]{\if0{#1}\fi}
 \newcommand*{\delete}[1]{\if0{#1}\fi}
\else
 \newcommand*{\add}[1]{\textcolor{red}{#1}}
 \newcommand*{\id}[1]{\textcolor{red}{#1}}
 \newcommand*{\delete}[1]{\sout{#1}}
\fi

\ifx00  
 \newcommand*{\addr}[1]{{#1}}
 \newcommand*{\idr}[1]{\if0{#1}\fi}
 \newcommand*{\deleter}[1]{\if0{#1}\fi}
\else
 \newcommand*{\addr}[1]{\textcolor{red}{#1}}
 \newcommand*{\idr}[1]{\textcolor{red}{#1}}
 \newcommand*{\deleter}[1]{\sout{#1}}
\fi

\newcolumntype{d}[1]{D{.}{.}{#1}}

\journal{Integrated Computer-Aided Engineering}
\firstpage{1}
\lastpage{15}
\volume{}
\pubyear{2020}

\begin{document}

\begin{frontmatter} 
\title{\id{(\#42)}\delete{Intermediate Pattern Representation} \delete{for} \add{Image-based Textile Decoding} \delete{with Deep Neural Network} }
\runningtitle{Image-based Textile Decoding}

\author{\fnms{Siqiang} \snm{Chen}},  
\author{\fnms{Masahiro} \snm{Toyoura}\thanks{Publised in Integrated Computer-Aided Engineering. The final publication is available at IOS Press through http://dx.doi.org/10.3233/ICA-200647. Corresponding authors: Masahiro Toyoura,  Department of Computer Science and Engineering, University of Yamanashi, Takeda 4-3-11, Kofu, Yamanashi, Japan. E-mail: mtoyoura@yamanashi.ac.jp. Gang Xu, School of Computer Science and Technology, Hangzhou Dianzi University, Hangzhou 310018, Zhejiang, China. E-mail: gxu@hdu.edu.cn.}}, 
\author{\fnms{Takamasa} \snm{Terada}},  
\author{\fnms{Xiaoyang} \snm{Mao}},  
\author{\fnms{Gang} \snm{Xu}}
\runningauthor{S. Chen et al.}
%
\address{\id{(\#80)}\add{School of Computer Science and Technology, Hangzhou Dianzi University, Hangzhou 310018, Zhejiang, China}}
\address{\add{Department of Computer Science and Engineering, University of Yamanashi, Takeda 4-3-11, Kofu, Yamanashi, Japan}}


\begin{abstract}
\id{(\#1, \#6)}
\add{A textile fabric consists of countless parallel vertical yarns (warps) and horizontal yarns (wefts). While common looms can weave repetitive patterns, Jacquard looms can weave the patterns without repetition restrictions. A pattern in which the warps and wefts cross on a grid is defined in a binary matrix. The binary matrix can define which warp and weft is on top at each grid point of the Jacquard fabric. }
\delete{A binary matrix pattern can define which warp or weft is on top at each grid point of a textile, and a Jacquard loom can accept the pattern to weave the textile fabric.}
The process can be regarded as {\it encoding} from pattern to textile. In this work, we propose a {\it decoding} method that generates a binary pattern from a textile fabric that has been already woven. We could not use a deep neural network to learn the process based solely on the training set of patterns and observed fabric images. The crossing points in the observed image were not completely located on the grid points, so it was difficult to take a direct correspondence between the fabric images and the pattern represented by the matrix in the framework of deep learning. 
Therefore, we propose a method that can apply the framework of deep learning via the {\it intermediate representation} of patterns and images. We show how to convert a pattern into an intermediate representation and how to reconvert the output into a pattern and confirm its effectiveness. 
In this experiment, we confirmed that $93\%$ of correct pattern was obtained by decoding the pattern from the actual fabric images and weaving them again.
\end{abstract}

\begin{keyword}
Textile\sep fabrication\sep intermediate representation\sep pattern decoding\sep \id{(\#1)}\add{Jacquard}\delete{Jacqurd} fabric
\end{keyword}

\end{frontmatter}

\section{Introduction}
With the rapid development of modern production techniques, the textile industry has undergone rapid changes. Intelligent weave machines have been introduced to make textiles convenient to produce and diversified in style and pattern. In addition, we have a clear need for personalized customization and on the whole, are not satisfied with traditional, uniform styles. Many of the ancient textiles exist as real objects only, and the patterns are still valuable and not outdated. If we want to reproduce such ancient fabrics, we need to analyze the patterns. Technicians can analyze the pattern by observing the fabric using a microscope and recording the crossing state of each yarn by disassembling them, but this is time consuming and tedious. At the same time, the original textile will be destroyed. Therefore, it is not a good way to solve this problem and we need to find a novel technique for automatically extracting the pattern. This would be useful for both reproduction and generating new patterns.

In this work, we focused on Jacquard fabric in which the crossing of warp and weft was specified by a binary matrix pattern. \id{(\#5, \#35)}\add{
In contrast to ordinary looms, which only weave repetitive patterns, Jacquard looms can weave free patterns without repetition restrictions. With Jacquard fabric, the warp and weft can be defined for each crossing point; the resulting fabric is not composed of uniform pattern regions. If the fabric is made with small repetitive patterns, the analysis can be done manually without much effort, but for Jacquard fabrics, the analysis of large patterns that make up the entire fabric is required, and manual analysis requires much effort. The segmentation based on pattern uniformity does not work for analyzing Jacquard textile patterns. }

\add{Jacquard textiles are often used in valuable fabrics such as traditional costumes and neckties, and there is high industrial value for the encoding and decoding of the patterns. Therefore, in this work, we focused on Jacquard fabric, in which the crossing of the warp and weft was specified by a binary matrix pattern.} 

The warp and weft yarns were dyed different colors and the pattern defined how the yarns crossed. The binary matrix pattern was defined as a binary image. For example, when the black yarn was above the white yarn, the crossing point was displayed in either black or white. We input the binary pattern into the weaving machine to produce the fabric. The Jacquard loom was able to accept the over-under relationship for the individual crossing points. 
\id{(\#39)}\add{Modern Jacquard looms can weave a coded pattern by loading a file of the pattern \cite{JacquardLoom}.}

\begin{figure}[ht]
\centering
\includegraphics[width=0.4\textwidth]{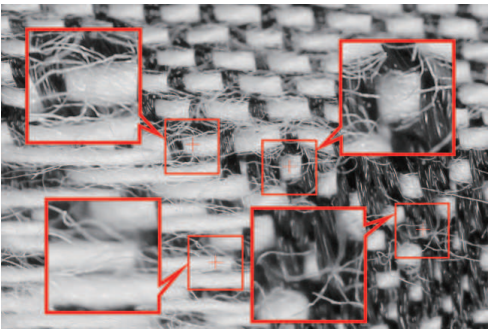}
\caption{\id{(\#33, \#44, \#70)}\add{The variety of crossing points makes the problem difficult. Conventional template matching cannot deal with the crossing points.}}
\label{fig:appearance}
\end{figure}

\id{(\#33, \#44, \#70)}\add{The appearances of images are different from each other, even at the same crossing point where a weft overlaps a warp. Fig. \ref{fig:appearance} shows zoomed-in images of the crossing point area in an observational image. Conventional template matching was not able to estimate the exact positions of the crossing points.}

Where there are many pairs of fabric images and their corresponding binary patterns, deep neural networks (DNNs) can {\it decode} unknown binary patterns from a corresponding fabric image. However, DNNs cannot directly output a binary pattern with thousands of crossing points from a fabric image with millions of pixels. To solve the problem, we introduced an {\it intermediate representation} that bridged the pair and enabled us to output the intermediate representational patterns by a DNN. In addition, to convert a fabric image into the intermediate representation, and the intermediate representation into a binary pattern, we built a practical method for the conversion. 
\id{(\#45,\#79)}\add{We introduced intermediate representation images because we expected the images to directly represent the likelihoods which are the crossing points. In addition, due to insufficient sample size, complex networks that require a large amount of training data are not likely to be trained well. A more complex network, like an end-to-end network which directly answers the positions of crossing points, trained by numerous samples could provide accurate results.}

\id{(\#12,\#21,\#30, \#76)}\add{We believe that the objective of this study, textile pattern decoding, has not been done to the best of the authors' knowledge in previous work. Therefore, we have not found any previous method that allows direct comparison of accuracy. Instead, by changing the parameters and kinds of filters in various ways, we confirmed the best possible settings for the current situation. } 

\begin{figure*}[ht]
\centering
\includegraphics[width=\textwidth]{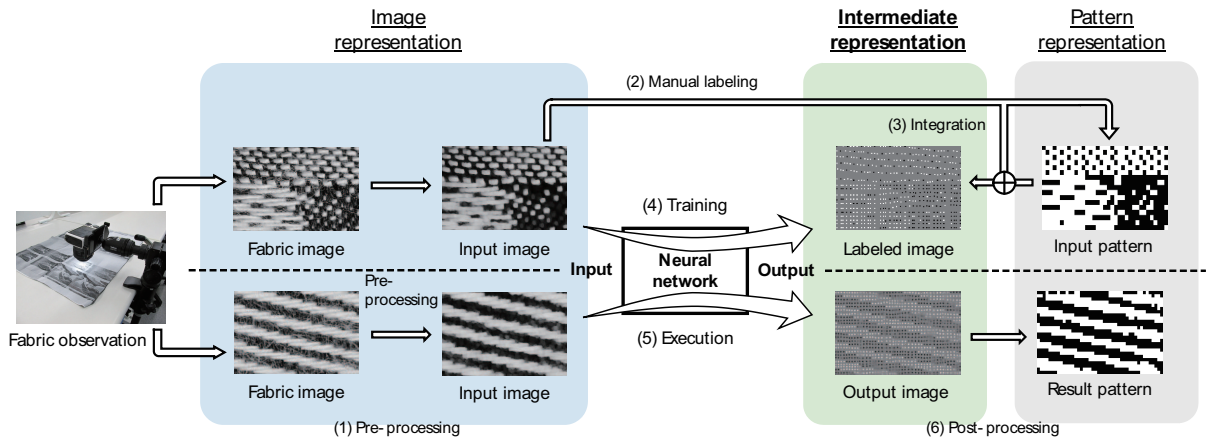}
\caption{The overview of our proposed method including the steps of (1) Pre-processing; (2) Manual labeling; (3) Integration; (4) Training; (5) Execution; and (6) Post-processing.}
\label{fig:overview}
\end{figure*}

\id{(\#46)}The contributions of this paper are as follows: 

\begin{enumerate}
\item Introduction of an intermediate representation for textile decoding tasks;
\item \id{(\#11, \#41)}\add{Interface design for manual tagging of cross points on an observed textile image (Section 3.1.1);} 
\item Proposal of a pre-process for converting a fabric image to an intermediate representation (Section \add{3.1.3}\delete{III.A}); and
\item  Proposal of a post-process for converting an output intermediate representation pattern into a regular binary matrix pattern (Section \add{3.3}\delete{III.C}).
\end{enumerate}

Fig. \ref{fig:overview} shows the overview of our proposed method. Section \id{(\#61)}\add{2}\delete{II} introduces the background and related work, and Section \add{3}\delete{III} shows a method of converting a fabric image into an intermediate representation pattern as well as an intermediate representation pattern into a binary pattern. We also describe the details of DNN configurations. In Section \add{4}\delete{IV} we present experimental results and in Section \add{5}\delete{V}, a summary. 

\section{Related Work}
\subsection{Pattern Creation by Computer Support}
Depending on the local density of the crossing points as well as the colors of yarn appearing on the top, a textile pattern brings different levels of brightness. Inappropriate patterns result in misaligned grid points and partial fabric stiffness. Considering many conditions, and through trial and error, textile patterns have been manually created since ancient times. In recent years, systems for designing textile patterns with the support of computers have been proposed and it has become possible to create more complex patterns \cite{Igarashi17,Ng14,Zhang18}. 

Toyoura et al. \cite{Toyoura19} proposed a dithering method for reproducing smoothly changing tones and fine details of natural images on woven fabric, focusing on representing gray scale images by using two colors of warp and weft yarns. The weaving pattern is generated by binarizing the input image using dither masks. The step dithering method alternately places values of $0$ and $255$ at given intervals in each row of the dither mask such that there is at least one cross point of warp and weft yarns in the spacing in the resulting fabric. Within these intervals, the threshold increases from $0$ to $255$ or from $255$ to $0$. This forms a stepping structure up and down. The resulting binary image reproduces the brightness of the input image while limiting the number of crossing points in the weave pattern. 

By modeling and rendering 3D CG (computer graphics) from a textile pattern, the woven result can also be predicted. By adding a physical collision detection, it is possible to reproduce the appearance of fabrics by CG. Users will be able to modify the pattern without actually weaving it.  

There have been many studies on computer graphics to generate photo-realistic fabric images from defined patterns \cite{Dobashi19,Leaf14,Sadeghi13,Zhao16}. The images obtained by observing yarns, fabrics, and 3D yarn data obtained by CT are used to generate realistic images. 3D fabric models are constructed from the data. These are the opposite of directional studies, as our proposed method restores patterns from actual fabrics. \id{(\#24)}\delete{Realization of our proposed method means the actual fabric can be reproduced from 3D CG.}\add{In this work, we aimed to output a pattern of weaving, which would be a complementary work.}

\subsection{Pattern Analysis for Fabric Images}
\idr{(\#2)}Due to industrial demand, \id{(\#3,  \#68)}\add{the pattern analysis by DNNs has been introduced for content-based image retrieval \cite{Hamreras20}, }
\addr{noisy image recognition \cite{Koziarski17}, }
\add{3D medical image super-resolution \cite{Hemsi20}, }
\addr{video surveillance \cite{Benito20}, foreground detection \cite{Garcia20}, multi-object tracking \cite{Yang19}, } 
\add{explosive device detection \cite{Donnelly20}, } 
\addr{pupil detection \cite{Vera19}, online data streaming \cite{Lara20}, airport baggage handling \cite{Sorensen20}, }
\add{and many other objectives. } 
\addr{Computational photography \cite{Wu19} and purpose-specific machine learning \cite{Rostami17} have been also employed to solve industrial problems. }

Many studies have been conducted to detect defects from fabric images. Huangpeng et al. \cite{Huangpeng18} constructed a weighted, low-rank, representational model of textures and detected defects. Ren et al. \cite{Ren17} realized the detection from a small sample set using pre-trained \delete{deep neural networks} \add{DNNs}, and Wei et al. \cite{Wei19} also employed DNN for \add{the} classification of fabric \delete{defect} \add{defects} \delete{only from the} \add{using a} small \delete{size} \add{number} of samples. 
\add{Jeyaraj et al. \cite{Jeyaraj19} improved the accuracy of defect detection by introducing the characteristic of texture in the training of DNN. Jing et al. \cite{Jing19} analyzed the detailed parameter settings of pre-trained networks, image patch sizes, the number of layers, and so on for detecting defects in repetitive patterns. }
Li et al. \cite{Li16} proposed a method for detecting defects by Fisher criterion-based stacked denoising autoencoders. Liu et al. focused on information entropy and frequency domain saliency \cite{Liu20}. 
\add{The ability to detect defects is valuable for the industry} \addr{\cite{AItexWeb,Tirronen08,Kwak01}}
\add{and patents for defect detection have been published \cite{Patent01}. On the other hand, }
defect detection is a task of anomaly detection or classification; however the pattern analysis of interest in this work is the binarization of the regularly aligned crossing points of warp and weft yarns. 

In order to measure the density of the weft for the woven fabric, Schneider et al. detected the intensity minima of brightness and estimated the positions for each \cite{Schneider14}. Compared with the simple Fourier transform, the change in the density of the weft can be obtained more accurately. Luo and Li \cite{Luo20} employed image processing techniques for quantifying fiber distribution uniformity in blended yarns based on cell counting and dilation area estimation. Meng et al. \cite{Meng19} realized their goal of estimating the yarn density of multi-color woven fabrics by deep learning. In this method, assuming that the whole fabric is woven in \id{(\#25)}\add{a periodical}\delete{the same} pattern, the intersection of the warp and the weft arranged on a regular grid is detected. 
\id{(\#73)}\add{Near-infrared spectroscopy revealed the material of yarns \cite{Sun15}. Machine learning also contributed to estimate the kinematic behavior of yarns \cite{Ribeiro20}.} 
In this paper, we aimed to analyze different patterns in a fabric. The grid formed by the warp and the weft may be greatly collapsed. 

Zheng et al. \cite{Zheng19} proposed a method for recognizing how predefined patterns are arranged on a piece of fabric. 
\id{(\#23)}\add{Loke and Cheong used co-occurrence matrices for the recognition \cite{Loke09}. These methods} 
\delete{This method} can be applied when the fabric is composed of a small number of known patterns, but this cannot be assumed for woven fabrics in which the pattern is created manually. Therefore, we aimed to determine the intersection of each point on the grid. 

\vspace{0.5cm}

\noindent \id{(\#9)}\add{3. Textile Pattern Decoding from Observed Image} \\ 
\delete{3. Method}
\setcounter{section}{3}
\setcounter{subsection}{0}

\id{(\#10, \#62)}\add{Here, we describe a method for decoding binary textile patterns from observed images. The method consists of three parts; pre-processing for converting a fabric image to an intermediate representation (Section 3.1); using a DNN for generating a label image (Section 3.2); and post-processing for converting the intermediate representation into a binary matrix textile pattern (Section 3.3). The training data is a set of observed images and a corresponding set of intermediate representations of these images with manually labeled cross point positions. At runtime, the final binary pattern is output from an observed image only. Fig. \ref{fig:overview} shows an overview of the process.}
\delete{We employed deep learning and image processing techniques. As shown in the overview in Fig. \ref{fig:overview}, a deep neural network outputs an intermediate representation image from an input image and post-processes the image to obtain the final binary pattern. We manually created intermediate representation images for training the deep neural network. }

\subsection{Pre-process for converting a fabric image to an intermediate representation}

\add{\subsubsection{Manual tagging of crossing points}} 

 \begin{figure*}[ht]
\centering
\begin{tabular}{c c c}
 \includegraphics[width=0.225\textwidth]{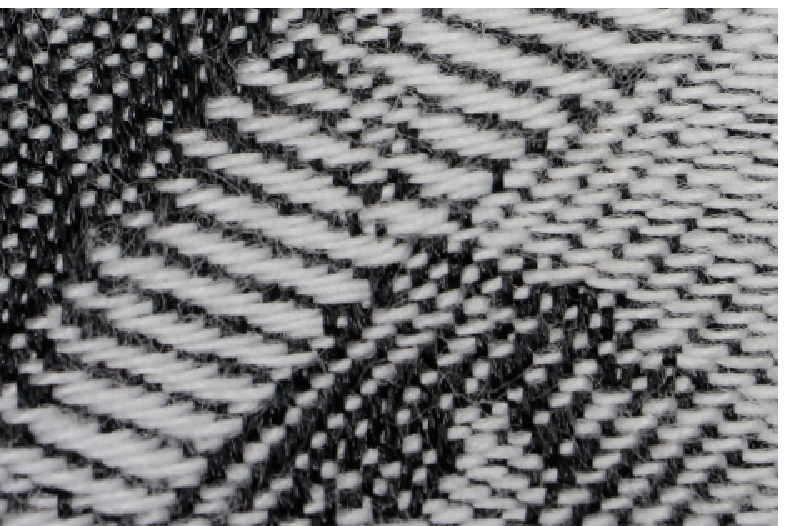}
 &
 \includegraphics[width=0.25\textwidth]{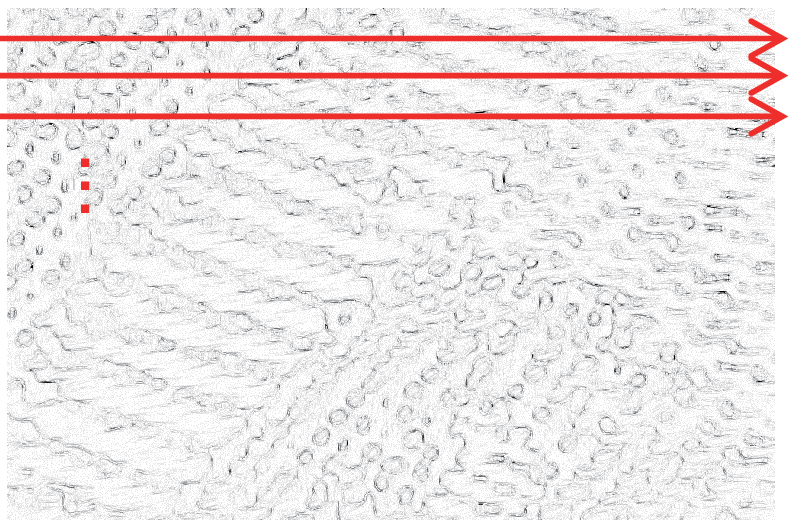}
 &
 \includegraphics[width=0.20\textwidth]{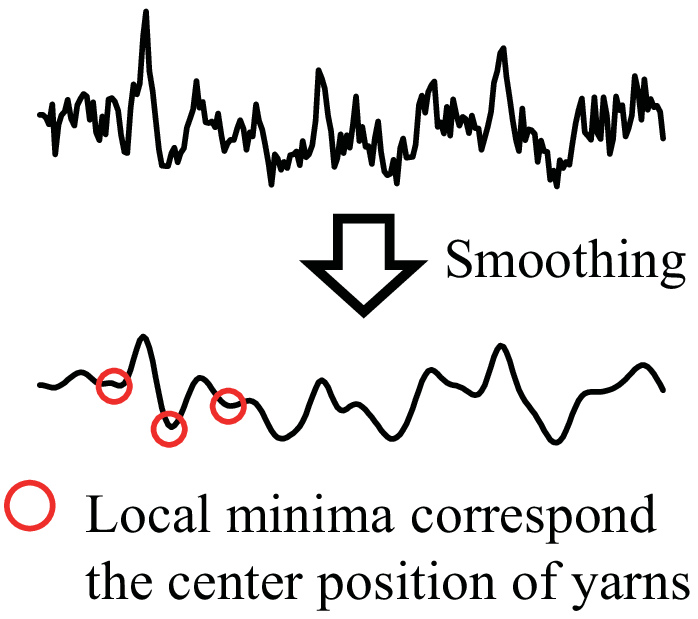}
 \\ 
 \add{\small (a) Observed image} 
 & 
 \add{\small (b) LOG filtered image}  
 & 
 \add{\small (c) Yarn position estimation} 
\end{tabular}
\caption{Estimation of warp and weft positions. \id{(\#15, \#56, \#63)}\add{For the input image shown in (a), the LOG filter image shown in (b) is calculated ($\sigma$=11 for Gaussian filter). In order to find the position of wefts, the sum of the pixel values in the LOG image is calculated row by row. (c) For a 1D sequence of sums, we smooth the values by finding the average of a certain value and the five neighboring values before and after the value. The local minima of the sequence indicate the positions where the power of edges is low, so they are output as the center positions of wefts. To estimate the positions of warps, the local minima of the sequence of vertical sums.}
\delete{The obtained image is filtered by a LOG filter. The row or column whose integral of edge intensity reaches a local minimum is likely to correspond to the center positions.}}
\label{fig:LOG}
\end{figure*}

Analysis of the weaving pattern can be done manually by observing the fabric using a microscope and recording the cross state of each yarn by disassembling it. This method is very time consuming, laborious, inefficient, and costly. In addition, it is also undesirable to perform such destructive inspections on textiles of high historical value. As is shown in Fig. \ref{fig:LOG}, we estimated fabric patterns from captured images and interactively modified fabric patterns to analyze weaving patterns.
Initial estimation results of the crossing point positions of warp and weft yarns were given by image analysis. The positions were interactively modified with a GUI so that the pattern could be obtained in a short time without destroying the actual fabric. Fig. \ref{fig:LOG} outlines the detection of the positions of weft yarns by filtering the image with a Laplacian of Gaussian (LOG) filter. \id{(\#14,\#27)}\add{A LOG filter is used to extract edges, such as regional boundaries, while removing small noise. We expected to ignore the fine threads and to output the boundaries between warp and weft yarns. Where the boundary pixels are few, it can be assumed that it is the center of the crossing point of warp and weft yarns.} 
Since \add{the} edge pixels could be observed at the edge of the yarn, the center of the yarn was found by picking up the local minima with fewer edge pixels \cite{Zheng19}. 

\begin{figure*}[ht]
\centering
\begin{tabular}{c c c c}
  \includegraphics[width=0.20\textwidth]{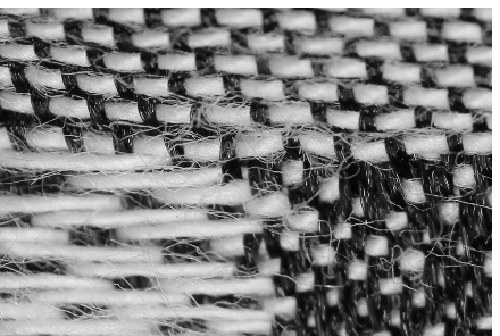}
  & \includegraphics[width=0.20\textwidth]{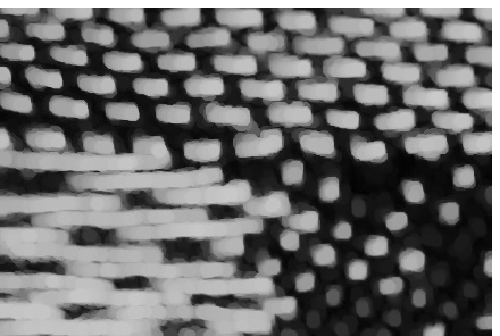}
  & \includegraphics[width=0.20\textwidth]{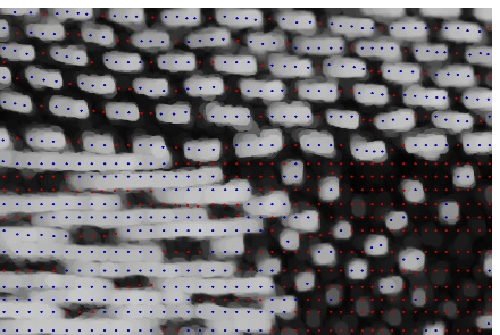}
  & \includegraphics[width=0.20\textwidth]{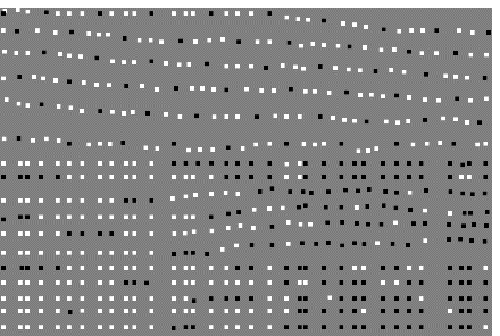}
    \\
  {\small (a) Observed image.} 
  & {\small (b) Pre-processed image.} 
  & {\small (c) Manually tagged crossing } 
  & {\small (d) Filtered labeled image.} 
  \\ 
  & 
  & \multicolumn{1}{l}{{\small \ \ \ points.}}
  & 
\end{tabular}
\caption{Converting processes of input image into labeled image. The observed image was captured with a camera with a macro lens. The image was smoothed in the pre-process and the positions of the crossing points were manually tagged. By filtering the crossing points, we obtained the final, labeled image.}
\label{fig:creation}
\end{figure*}

When the pixel values were integrated in the horizontal direction (X direction), the intensity of the edge component in each row was obtained. The position where the edge intensity reached the maximum was the end of the row with many edge elements, and the row whose edge intensity reached the local minimum was the center of the yarn with relatively few edges. Applying the same process to the warp yarns, the crossing points of the warp and weft yarns could be obtained as the initial positions of crossing points. Although this method can analyze patterns quickly, it is not automated enough to perform large-scale analysis. Therefore, we used this method only to prepare training data.

\id{(\#11, \#62)}\add{Once the initial positions of the warps and wefts were determined from the input image, we computed which of the warps or wefts was higher for each of the grid points. By giving two representative colors of warp and weft, we could see which color was closer at a grid point, and then give the state of the grid at that point as 0 or 1.}

\add{In the developed GUI, a user can add, remove, and move yarns by clicking and dragging on the screen. The operational mode was defined as move, add, or delete by key presses: The operation with the shift key pressed was for the warp; the operation with the ctrl key pressed was for the weft, and the operation without pressing any key was for the crossing point. When the weft and warp positions were updated, the initial intersection (based on the new warp and weft positions) could be obtained by pressing the C key. When the F key was pressed, the vertical direction of the weft and warp at the crossing point closest to the clicked mouse position was reversed.}

\add{\subsubsection{Noise removal of fine fibers}} 
In the observed image, even fine fibers are observed, as shown in Fig. \ref{fig:creation}(a). The fine fibers produce intensity edges which result in high values in the LOG-filtered image. This problem can be easily solved by the image processing of erosion and \id{(\#28)}\add{dilation}\delete{expansion}, resulting in the image shown in Fig. \ref{fig:creation}(b).
\id{(\#16)}\add{Erosion and dilation are used to remove the regions with little area. In our experiments, we used three functions in MATLAB; strel, imerode, and imdilate. To obtain the results, the image was eroded with a $5.5$ pixel radius setting and then expanded with the same setting. The radius setting was determined by checking to see if the thin fibers disappeared in the sample image.}

Fig. \ref{fig:creation}(c) shows the image with the manually labeled crossing points. The red and blue points are central with the red point indicating that the warp is on top at the point. The blue point indicates that the weft is on top at the point. Fig. \ref{fig:creation}(d) shows the image with white pixels for the point of the warp on the weft; black pixels for the point of weft on the warp; and gray pixels for other than the crossing points in the image. For machine learning, the pre-processed image in Fig. \ref{fig:creation}(b) and the intermediate representation image in Fig. \ref{fig:creation}(d) are provided as a pair; \id{(\#28)}\add{the} DNN receives the pre-processed image and outputs an intermediate representational image.

\begin{figure*}[ht]
\centering
\begin{tabular}{l l l l}
  \includegraphics[width=0.20\textwidth]{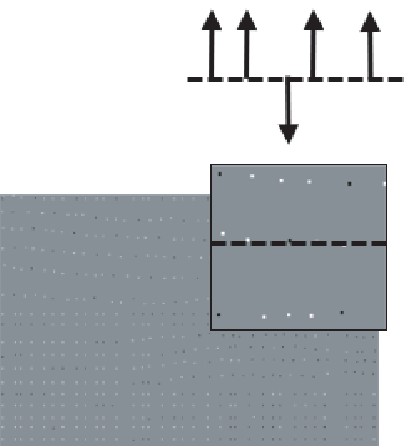}
  & \includegraphics[width=0.20\textwidth]{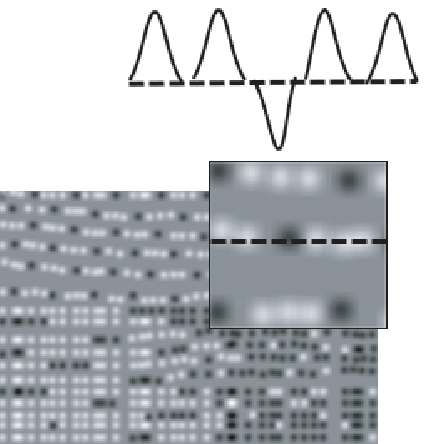}
  & \includegraphics[width=0.20\textwidth]{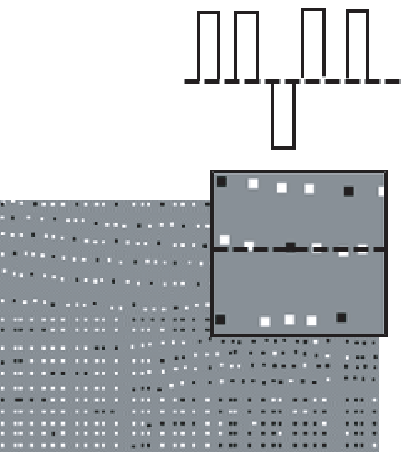}
  & \includegraphics[width=0.20\textwidth]{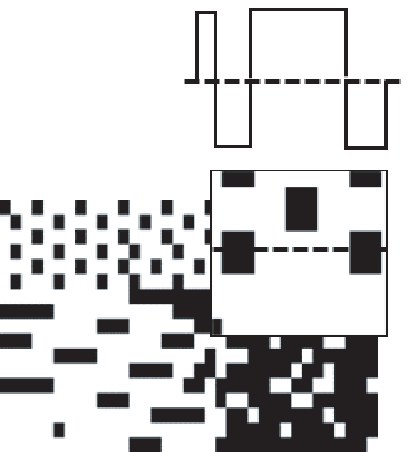}
  \\
  \multicolumn{1}{l}{\footnotesize (a) Impulse peak pattern.} 
  & 
  \multicolumn{1}{l}{\footnotesize (b) Gaussian-filtered peak } 
  & 
  \multicolumn{1}{l}{\footnotesize (c) Box-filtered peak pattern.} 
  & 
  \multicolumn{1}{l}{\footnotesize (d) Binary pattern.} 
  \\ 
  \multicolumn{1}{l}{\footnotesize \add{$I_0$, defined by Eq. (1). }} 
  & 
  \multicolumn{1}{l}{\footnotesize \add{pattern. $I_G$, defined by}} 
  & 
  \multicolumn{1}{l}{\footnotesize \add{$I_B$, defined by Eq. (3). }} 
  & 
  \\ 
  & 
  \multicolumn{1}{l}{\footnotesize \add{Eq. (2). }}
  & 
  & 
  \end{tabular}
\caption{\id{(\#17)}\add{Impulse peak pattern and three}\delete{Three} kinds of labeled images.}
\label{fig:three}
\end{figure*}

\add{\subsubsection{Converting intermediate representation images}}
Here, we considered how to make an intermediate representational image. If we were to use the result of manual labeling in its present form, we would get an image $I_0$ with impulses at the crossing points, as shown in Fig. \ref{fig:three}(a) \id{(\#17)}\add{and Eq. (1)}. At a crossing point, it would take a value of 0 or 1 and at the very next pixel, it would take a gray value of 0.5.

\begin{eqnarray}
  I_0(x,y) = 
  \left\{ \begin{array}{l l}
    1 & if \ weft \ on \ warp \ at \ (x,y), \\ 
    0 & if \ warp \ on \ weft \ at \ (x,y), \\ 
    0.5 & otherwise.  
  \end{array} \right.
\label{eqnarray:1}
\end{eqnarray}

In the input image, there is no significant change in the immediate vicinity of the crossing position, so it is expected that DNN would have difficulty producing such an image. The image $I_G$ with the Gaussian blur shown in Fig. \ref{fig:three}(b) can represent the situation in which the likelihood peaks at the crossing point and the likelihood gradually decrease. It is expected to be easier to handle with DNN than the impulses. $I_0$ can be reproduced by finding the minimum and maximum points of each peak. Furthermore, by trial and error, we found that the image with the box filter as shown in Fig. \ref{fig:three}(c) has better accuracy \id{(\#29)}\add{for} \delete{in} the resulting images. The results of using $I_0, I_G$ \id{(\#17)}\add{defined by Eq. (2)}, and $I_B$ \add{defined by Eq. (3)} for the training data, respectively, are presented in the Experiment section.
\begin{eqnarray}
  I_G &=& LOG(I_0), \\ 
  I_B(x,y) &=& \left\{ \begin{array}{l l}
    1 & \max_{(s,t) \in N(x,y)} I_0(s,t)=1, \\ 
    0 & \max_{(s,t) \in N(x,y)} I_0(s,t)=0, \\ 
    0.5 & otherwise.  
  \end{array} \right.
\end{eqnarray}

\noindent Note that $N(x,y)$ represents the set of neighborhood pixels of $(x,y)$. \id{(\#64)}\add{In Fig. \ref{fig:overview} and Fig. \ref{fig:creation}(b), we used a window of $9 \times 9$ pixels as $N(x,y)$. Different sizes of windows were also tried and examined in the experiment.} 

In Fig. \ref{fig:three}(d), the final binary pattern is transformed into an image, but the size of the image is much smaller than the input image. Additionally, the positions of the crossing points are different from those of the input image. To obtain such an image by DNN is difficult because DNN tries to determine whether a point is a crossing point based on the information of a pixel and its neighbors. In that situation, DNN cannot produce an image like Fig. \ref{fig:three}(d) in which the position is completely misaligned. 

\subsection{DNN model for generating label image}
We used a DNN to output a label image from the input image. It was necessary to determine whether a pixel corresponded to a crossing point by looking at the surroundings of the pixel without losing the accuracy of the position. We employed a DNN with a U-net structure \cite{Ronneberger15} to solve this task in which a network with this structure can take into account the surroundings by a network of autoencoders and maintain the resolution by jumping paths. 

\begin{figure}[ht]
\centering
\includegraphics[width=0.47\textwidth]{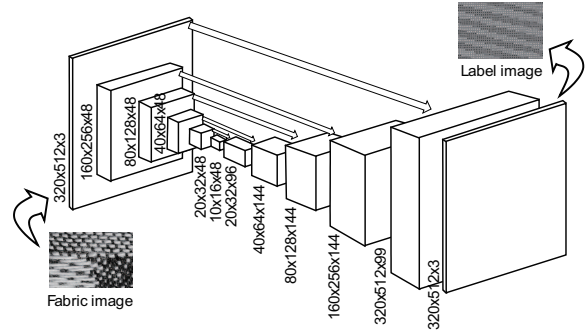}
\caption{Neural network model structure for converting labeled intermediate representation image.}
\label{fig:network}
\end{figure}

The model structure is shown in Fig. \ref{fig:network}. \add{\id{(\#49, \#53, \#54)} The input to the DNN is a pre-processed image, which is an image with fine noise removed, and the output from the DNN is an intermediate representation of the image, which shows the likelihood of the existence of the crossing points on the image.} The model has a total of six down-roll multi-layer and six upper convolution layers; the final output size is the same as the input size. When the pre-processed observation images were input, the network was trained so that the label images were output. \id{(\#55)}\add{The input image size of $320 \times 512$ was chosen because it is difficult to train a network with too many neurons due to the limited number of samples, and because it is easy to work with when tagging manually. It does not necessarily have to be this size; other sizes may be used with confidence.}

In the learning phase, we gave a pair of observed images and a label image that were manually generated. For the loss function, the L1 norm of pixel values was employed.
\id{(\#51)}\add{The number of neurons in each layer is also shown in the figure. PyTorch was used for implementation. A more detailed implementation environment is described at the beginning of the experimental section.}

\begin{figure}[ht]
\centering
\includegraphics[width=0.35\textwidth]{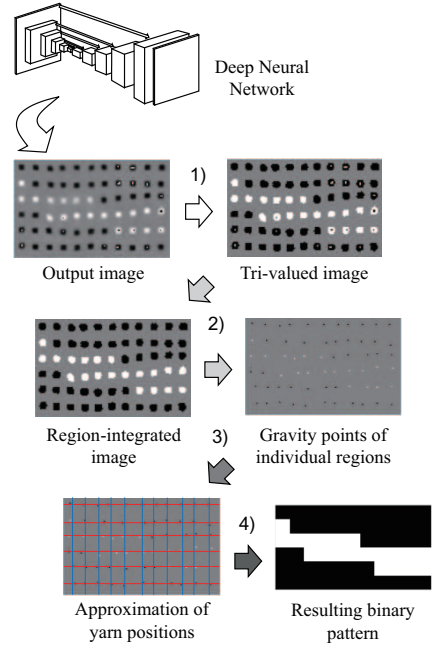}
\caption{Post processing for the conversion of output image from a deep neural network.}
\label{fig:transition}
\end{figure}

\subsection{Post-process for converting intermediate representation into binary matrix textile pattern}
DNN outputs an intermediate representation image; however, a non-trivial post-processing is required to obtain the final binary pattern. This occurs in several steps: 1) the intermediate representation image is converted into a tri-valued image of $0$, $0.5$, and $1$; 2) each peak region is merged into one; 3) the approximate horizontal and vertical positions of the warp and weft yarns are found; and 4) a determination is made as to which of warp and weft are over at each grid point. Finally, we get a binary pattern with 0s and 1s on the grid points, as shown in Fig. \ref{fig:transition}. 

\vspace{0.2cm}


\subsubsection{Converting the intermediate representation image into a tri-valued image}
The intermediate representation image \id{(\#66)}\add{output} \delete{outputted} from the DNN had continuous values in the range of $(0,1)$. In order to find the crossing points by segmenting the image, each pixel must have a discrete value. We replaced the value of each pixel with the closest of the three values, in which the crossing point of the warp was 0, the crossing point of the weft was 1, and 0.5 otherwise. This process is equivalent to thresholding by two values of 0.25 and 0.75. \id{(\#31)}\add{An image is converted into a tri-valued image only with white, black, and gray pixels.}

\vspace{0.2cm}

\subsubsection{Integrating regions of multiple values}

In the image obtained by \id{(\#31, \#52)}\add{the process described in  3.3.1}\delete{(1)}, based on the gray background, a region consisting of white pixels only, a region consisting of black pixels only, and a region containing both white and black pixels appeared. Since the crossing points were never adjacent to each other in real fabrics, no region ever contained both white and black pixels.

\id{(\#52)}\delete{Therefore,}\add{We aim to avoid the coexistence of white and black pixels in each region that represents a crossing point.} \delete{the}\add{The} region containing both white and black pixels was merged into the region with the larger number of white or black pixels.  \id{(\#31,\#38,\#52)}\add{The pseudo code for the process is described in Algorithm  \ref{Algorithm:CCL}.} 

{\small
\begin{figure}[ht]
\begin{algorithm}[H]
\caption{\id{(\#31,\#38,\#52)}\add{Integrating regions with multiple values with connected component labeling (CCL)}}\label{CCL}
\label{Algorithm:CCL}
\begin{algorithmic}[1]
\renewcommand{\algorithmicrequire}{\textbf{Input:}}
\renewcommand{\algorithmicensure}{\textbf{Output:}}
\algblockdefx[Foreach]{Foreach}{EndForeach}[1]{\textbf{foreach} #1 \textbf{do}}{\textbf{end foreach}}
\Require Tri-valued image 
\State Pixel value $v(p) \in \{0, 0.5, 1\}$ in tri-valued image 
\State 4-neighboring pixel of $p$ is $N(p)$
\State Function $BW(p,q)$ = True only when ($v(p)=0 \wedge   v(q)=1$) $\vee$ ($v(p)=1 \wedge v(q)=0$) 
\Ensure Tri-values image without 0-1 mixed regions 
\State \Comment{CCL segmentation}
\State CCL segments regions with unique values
\State Identify regions $\{r\}$ by IDs $\{i\}$ 
\State \Comment{Merge region IDs}
\State flag $\leftarrow$ True 
\While {flag = True}
\State flag $\leftarrow$ False 
\Foreach {Region $r_i$}
 \Foreach {Pixel $p \in r_i$} 
   \If {$\exists q \in N(p), BW(p,q) = True$} 
     \If {$p \in r_i$ and $q \in r_j$ \ and \ $i<j$}
       \State $j \leftarrow i$ 
       \State flag $\leftarrow$ True 
     \EndIf 
   \EndIf 
 \EndForeach
\EndForeach
\EndWhile
\State \Comment{Unify pixel values in individual regions}
\Foreach {Region $r_i$}
 \If {0-valued pixels are more than 1-valued pixels in $r_i$} 
  \State All pixels in $r_i$ $\leftarrow$ 0 
 \Else 
  \State All pixels in $r_i$ $\leftarrow$ 1 
 \EndIf
\EndForeach
\end{algorithmic}
\end{algorithm}
\end{figure}
}

The regions were labeled by segmentation and the adjacent regions for each region were searched. The connected components labeling (CCL) gave the individual, connected regions of a binary image drawn by identical numbers. A region containing both white and black pixels could be detected from the adjacency matrix. Ignoring the background gray region, we first flagged the adjacent regions. We then flagged the adjacent regions of the first set of adjacent regions and repeated the process until no more flags appeared. Finally, the black and white regions included in each independent region were recognized. The number of white and black pixels belonging to each region were counted and the region was merged by the higher color.

After we merged the regions with only white pixels and black pixels, we next found the representative points of each region. Since the representative point should be located in the center of the crossing point region, the pixel at the gravity point of each segmented region was to be the representative point of that region. 

\vspace{0.2cm}

\subsubsection{Finding approximate horizontal and vertical positions} 
Regional representative points indicating crossing points included duplicated and missing points. The crossing points were constraints that existed in a grid along the yarns. The regional representative points satisfying this constraint were picked up and converted into binary matrix patterns.

First, by detecting the positions of the warp and weft yarns, the approximate positions of the grid points were estimated. The distance transformation image for the obtained candidate crossing points were able to show the likelihood of a warp or weft passing through its position. The positions of warp and weft yarns were estimated in the same way as in the pre-processing, using the distance transformation image as the target.

The crossing point pixels were replaced by $1$ as the object region and the other pixels were replaced by $0$ as the background region. The value of each pixel in the distance transformation image indicated the distance from that pixel to the nearest pixel in the object region. The distance was $0$ for pixels in the object region; the further away the pixel was, the larger the value was.

We integrated the distance transformation values of pixels in a column. Along a column, the more pixels that were in the object region or close to it, the smaller the integral became while the likelihood of being the center of the warp or weft yarn increased. As described in Section 3.A, the positions of warp and weft yarns were estimated by determining the likelihood according to the power of intensity edges. The integral was smoothed in the same way to reduce the effect of local noise. By way of this process, the positions of the warp and weft yarns were estimated.

\vspace{0.2cm}

\subsubsection{Determining value at each grid point}
As the positions of the warp and weft yarns were estimated, the positions of the grid points were also estimated automatically. By assigning a value of $0$ or $1$ to each grid point, we were able to obtain the final binary pattern in a matrix form. In Fig. \ref{fig:transition}, the blue lines indicate the estimated location of the warps while the red lines indicate the estimated location of the wefts. The white and black points are representative points with the values of $1$ or $0$.

For each grid point, we extracted the best candidate point within distance $s$ and let the value of $0$ or $1$ of the candidate points be the value of the grid points. If there was more than one candidate point, we adopted the closest one to the grid point; if there was no point, we adopted the closer color of which the warp or weft colors from the color of the grid point itself. $s$ allowed for being set arbitrarily; a smaller value meant that we adopted only candidates strictly close to the grid point. In the experiment, we confirmed the accuracy of pattern reproduction according to the variety of $s$. 

\section{Experiment}

\begin{figure*}[ht]
\centering
\subfigure[Pre-processed input image.]{
\includegraphics[width=0.3\linewidth]{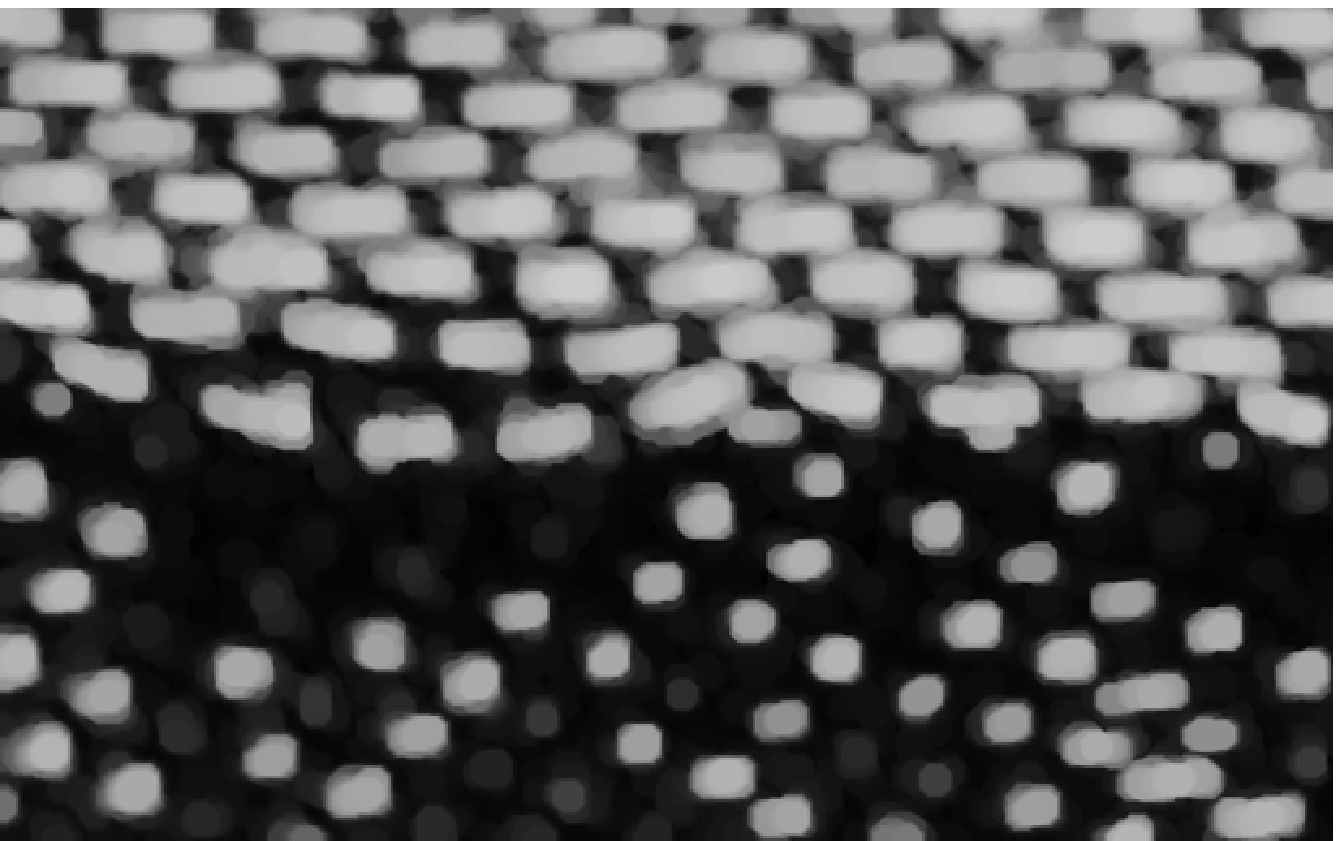}
\includegraphics[width=0.3\linewidth]{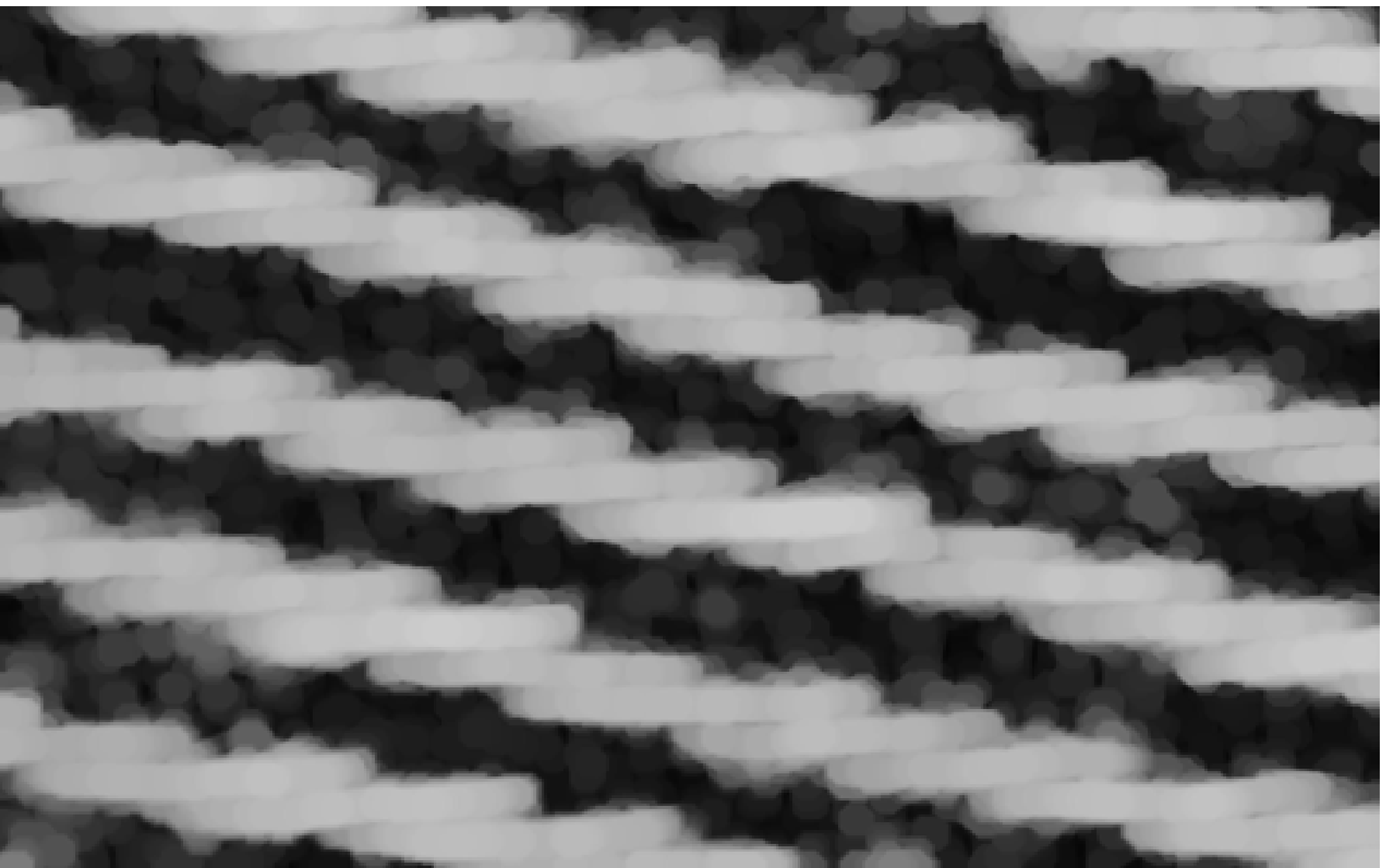}
\includegraphics[width=0.3\linewidth]{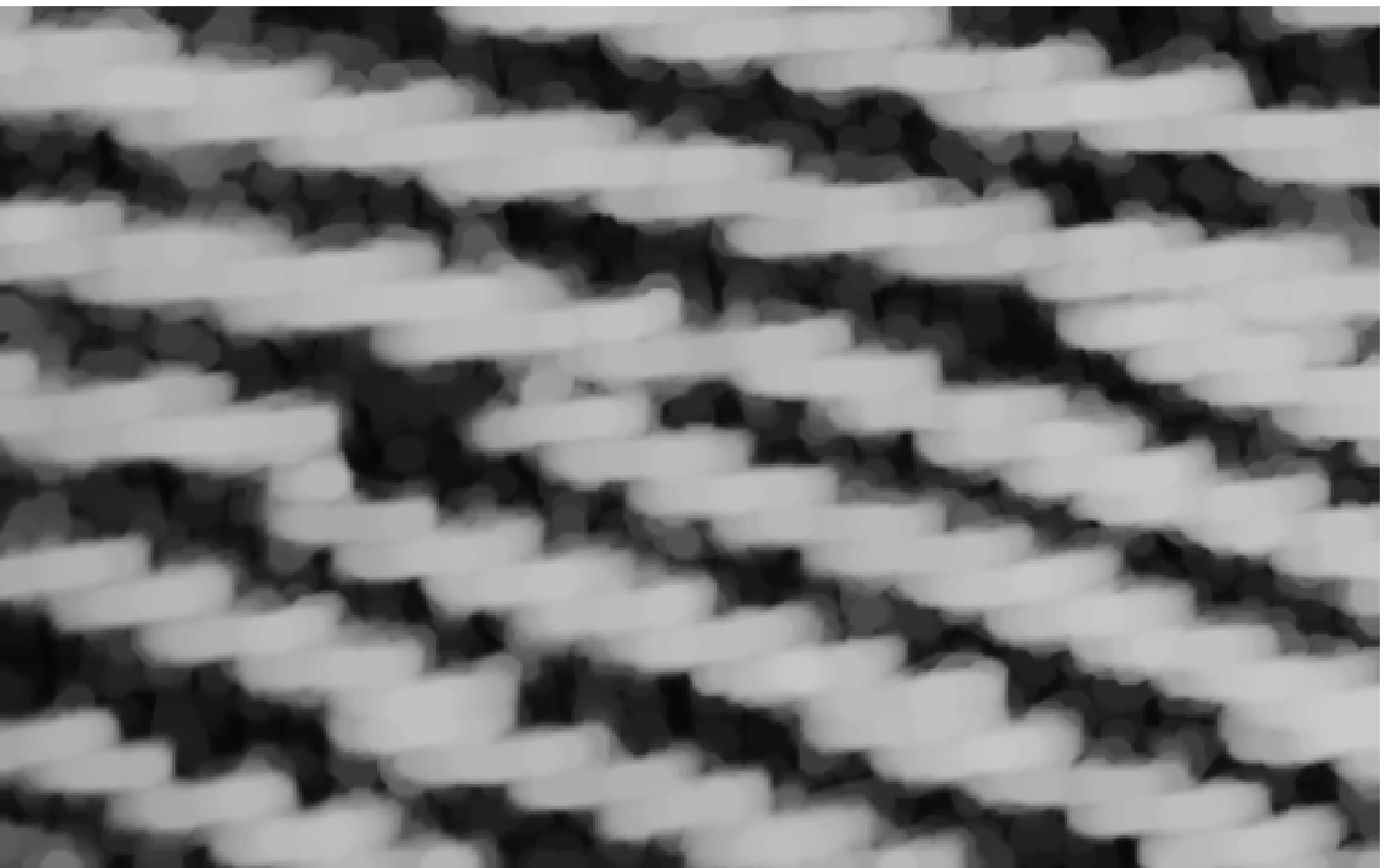}
}
\quad
\subfigure[Result image by impulse peak pattern training.]{
\includegraphics[width=0.3\linewidth]{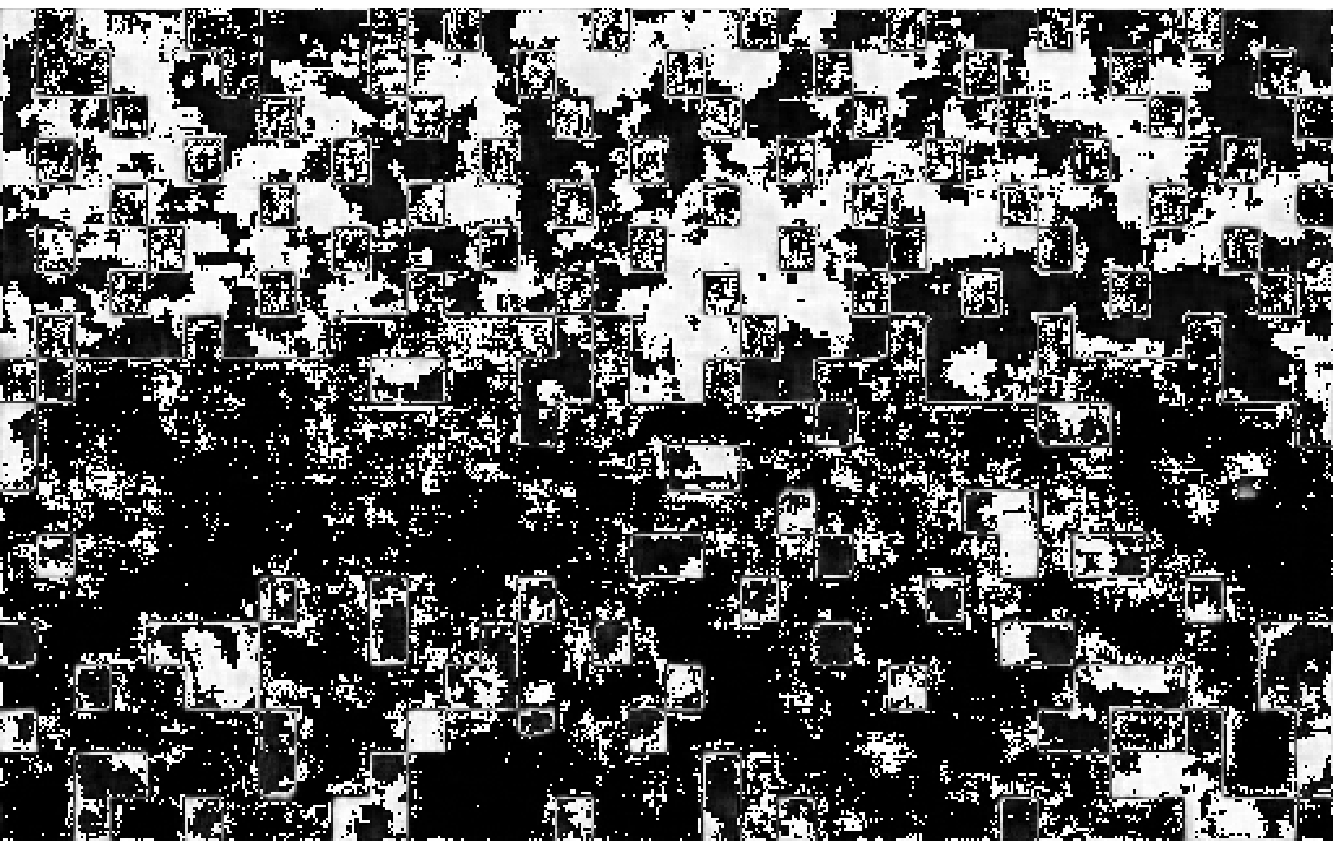}
\includegraphics[width=0.3\linewidth]{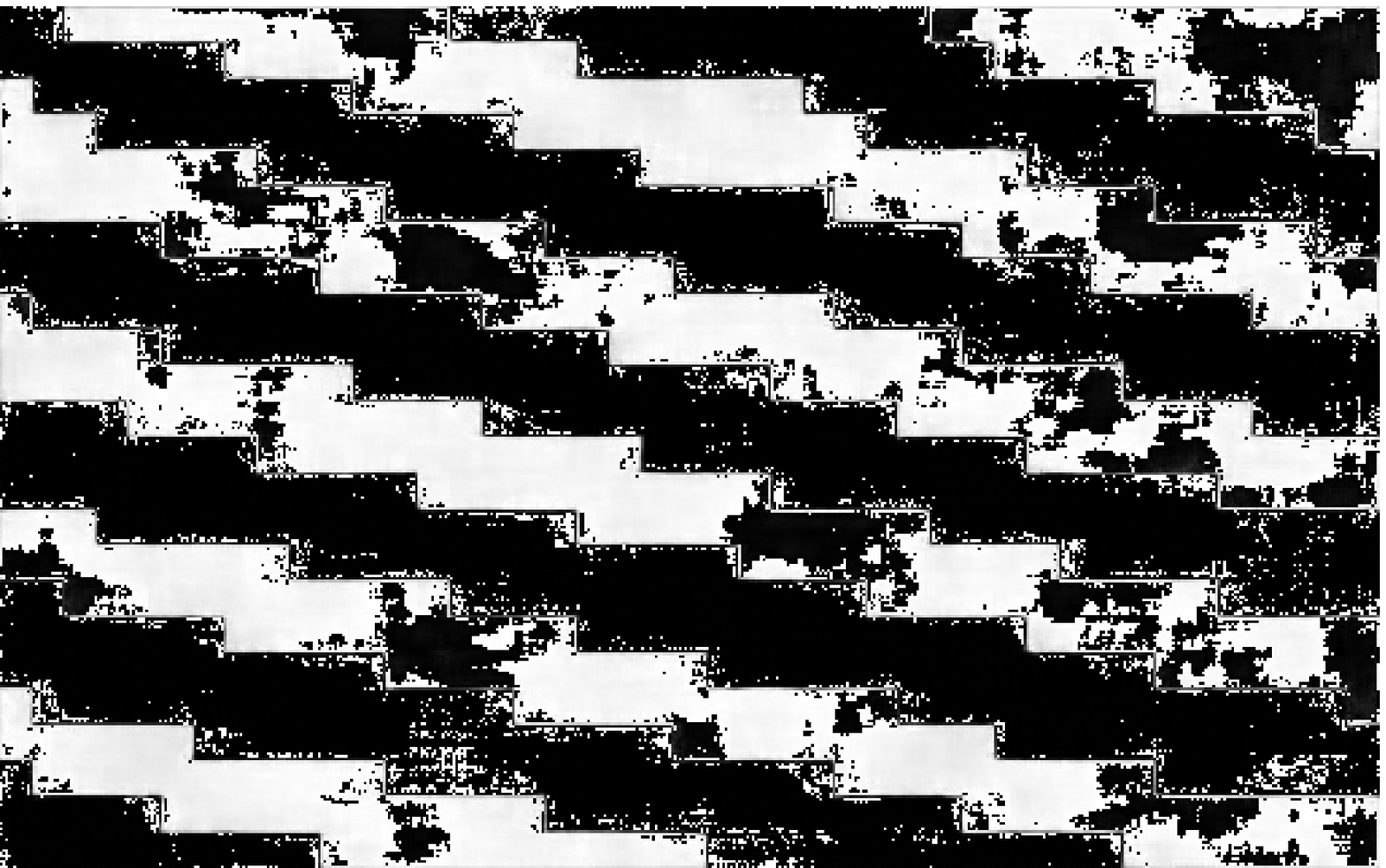}
\includegraphics[width=0.3\linewidth]{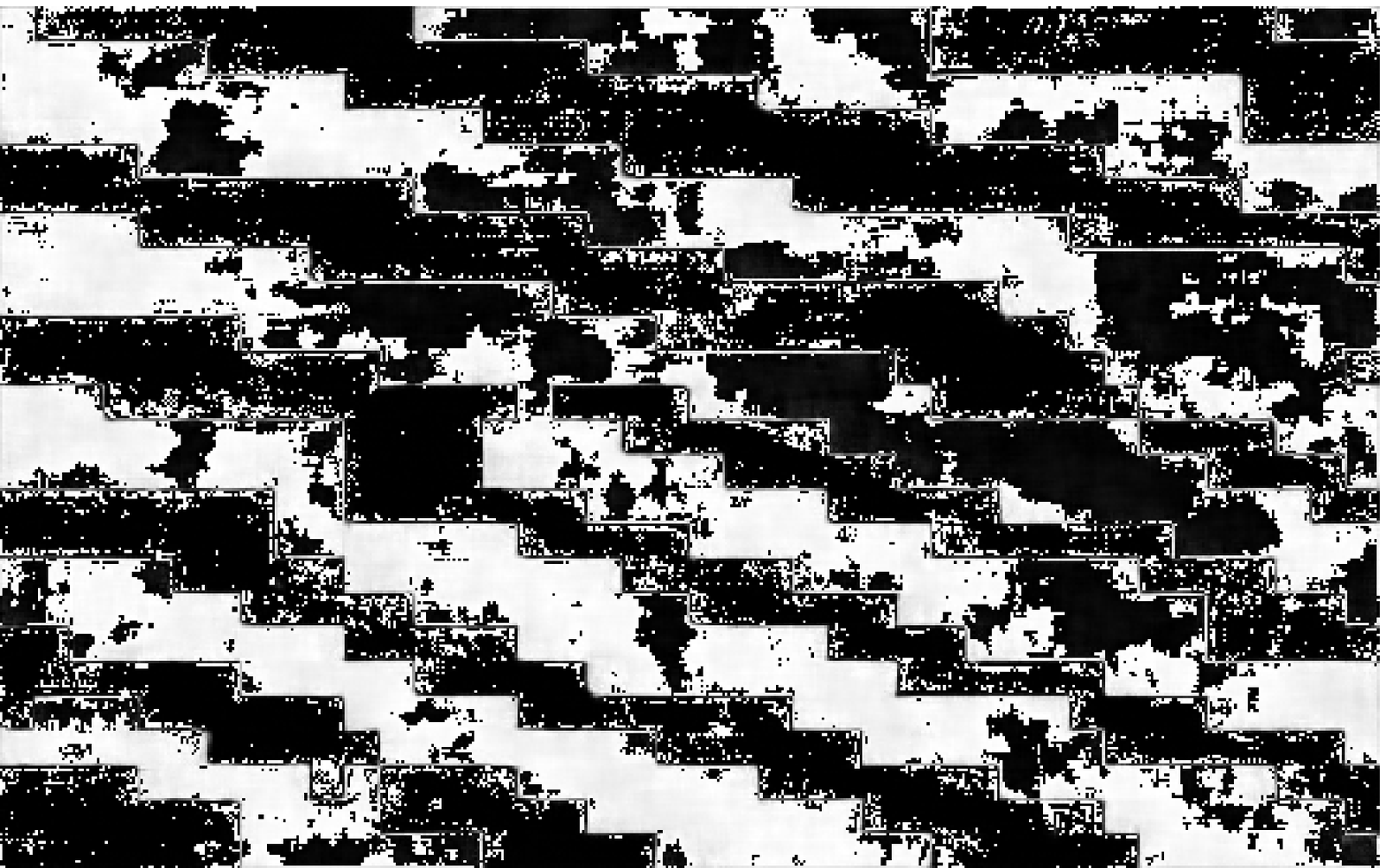}
}
\quad
\subfigure[Result image by Gaussian-filtered peak image training.]{
\includegraphics[width=0.3\linewidth]{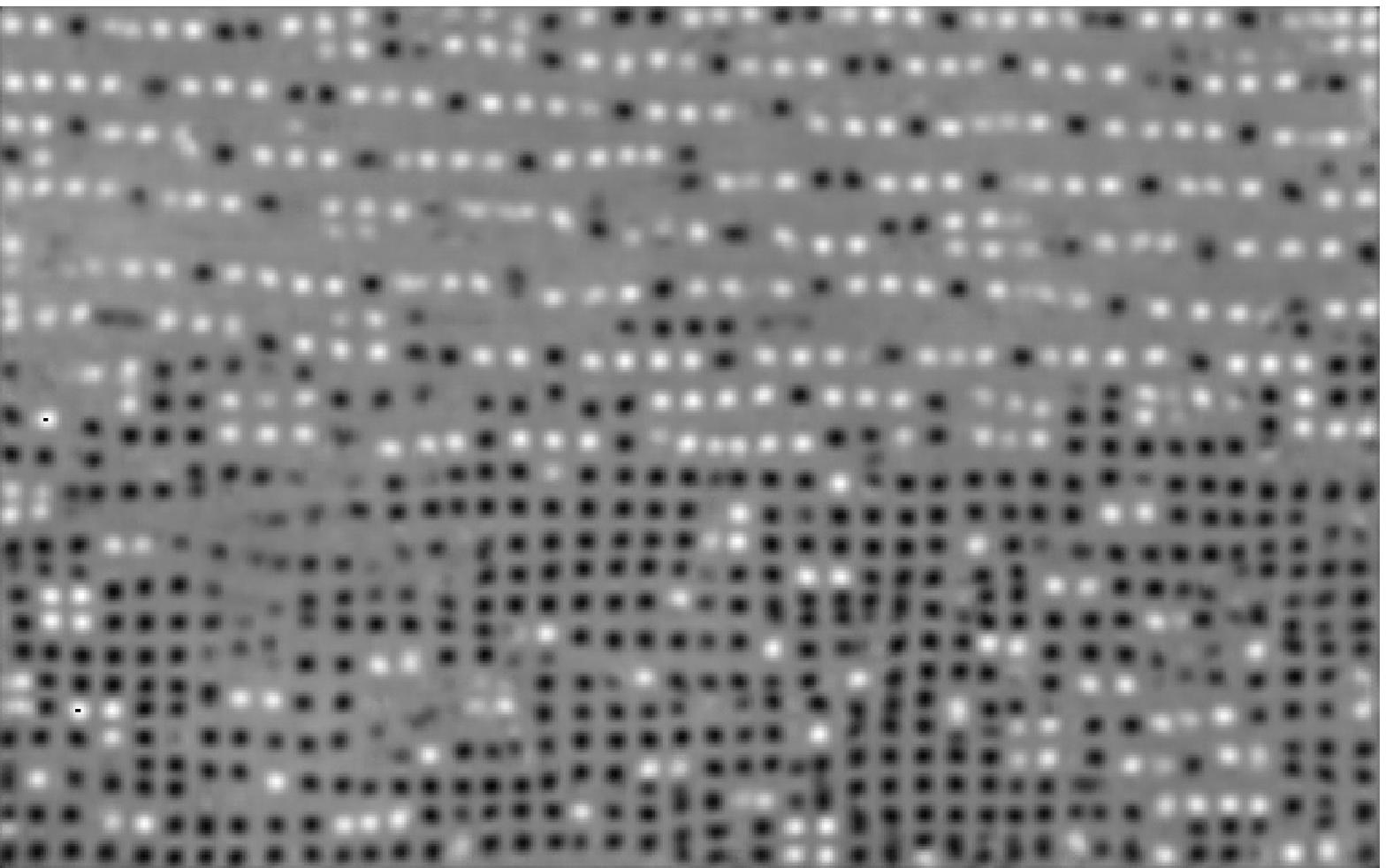}
\includegraphics[width=0.3\linewidth]{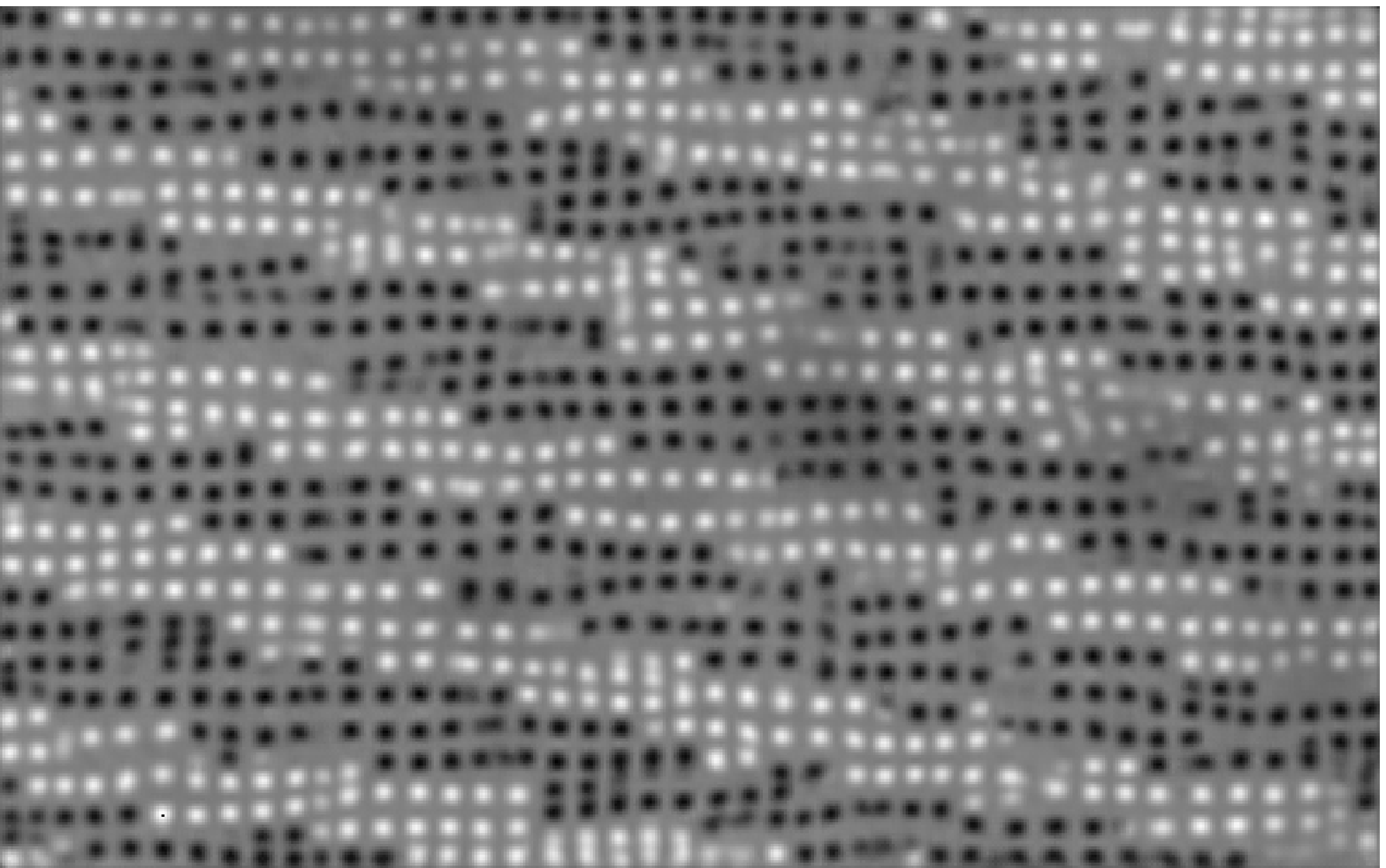}
\includegraphics[width=0.3\linewidth]{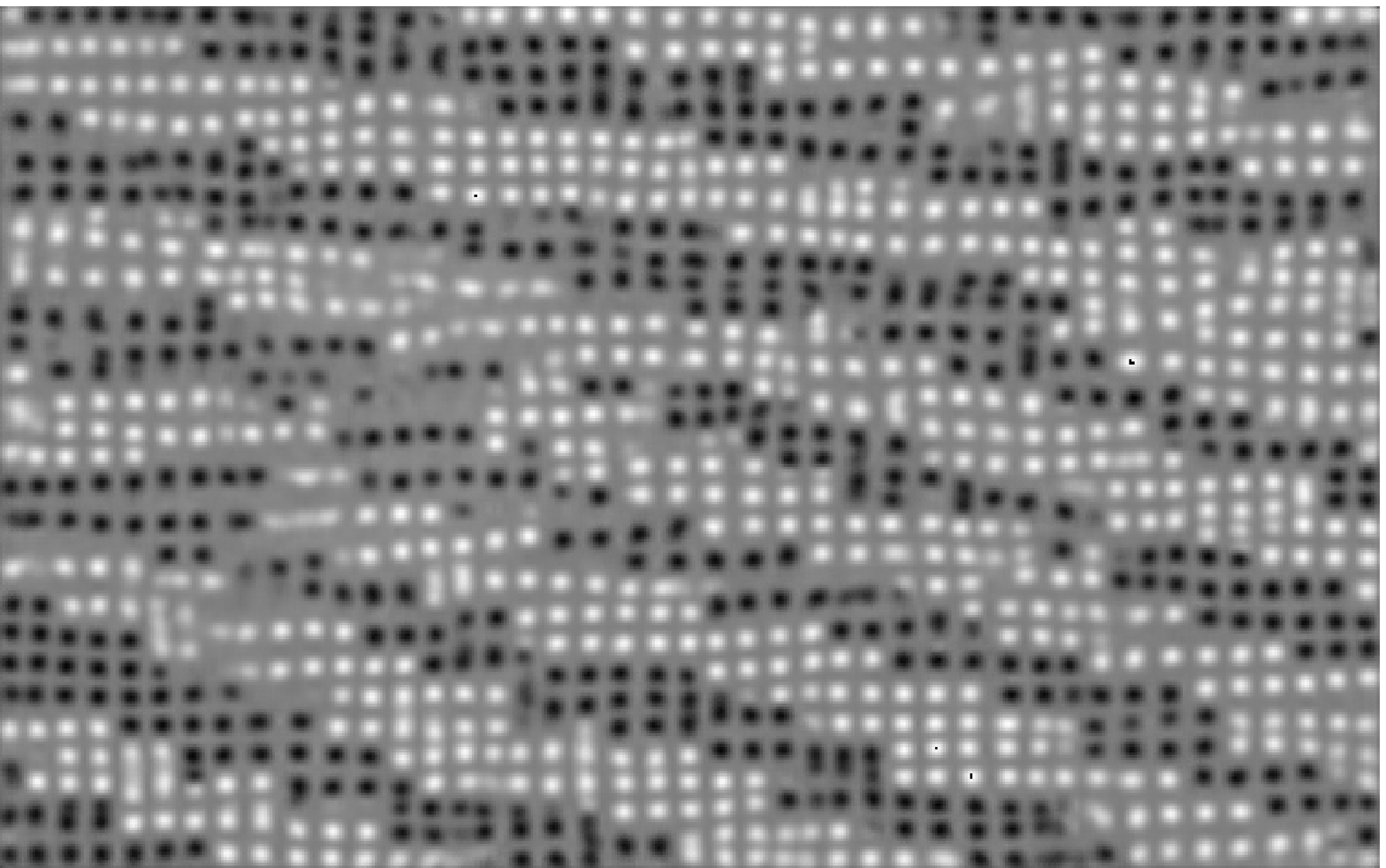}
}
\quad
\subfigure[Result image by box-filtered peak image training.]{
\includegraphics[width=0.3\linewidth]{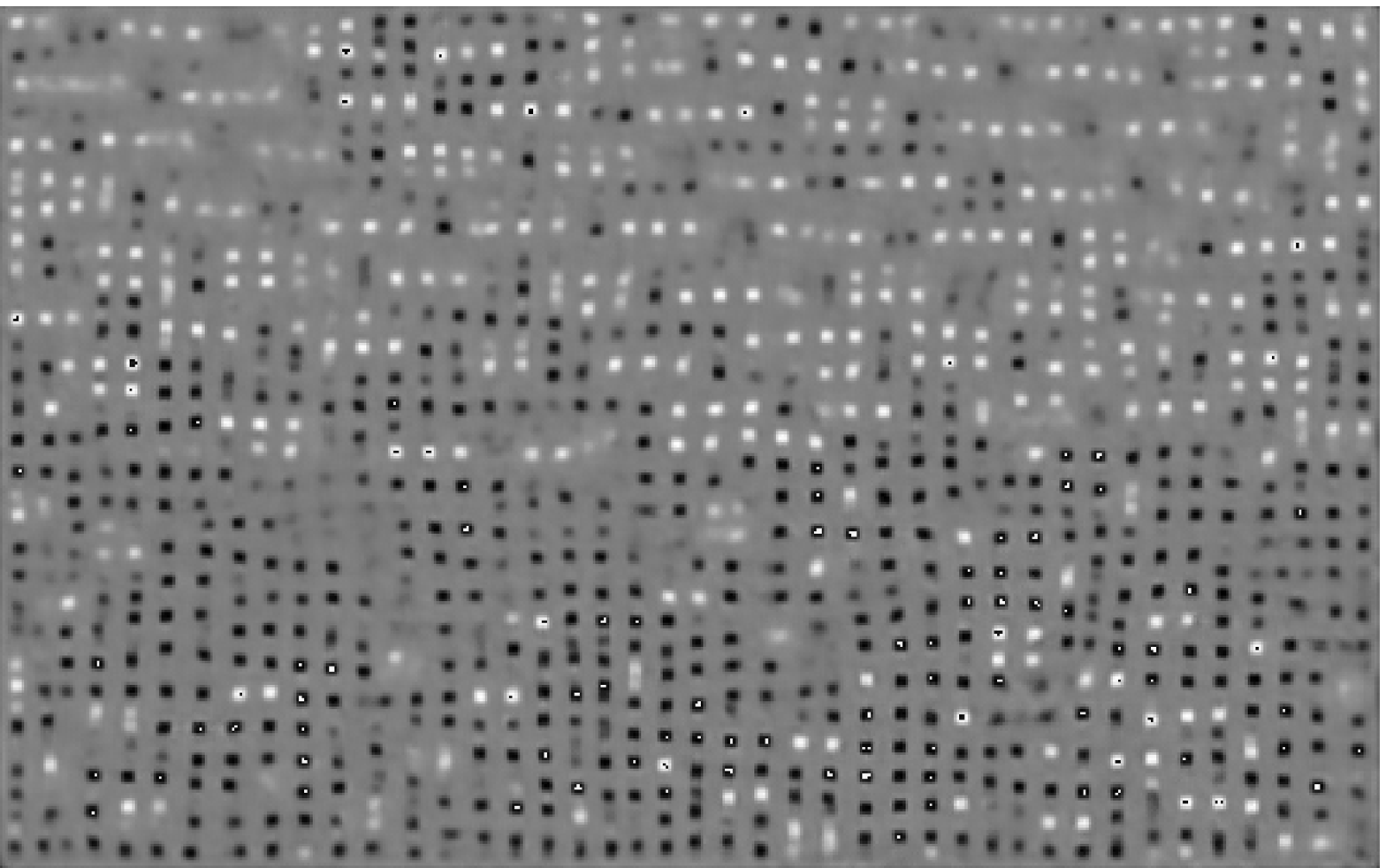}
\includegraphics[width=0.3\linewidth]{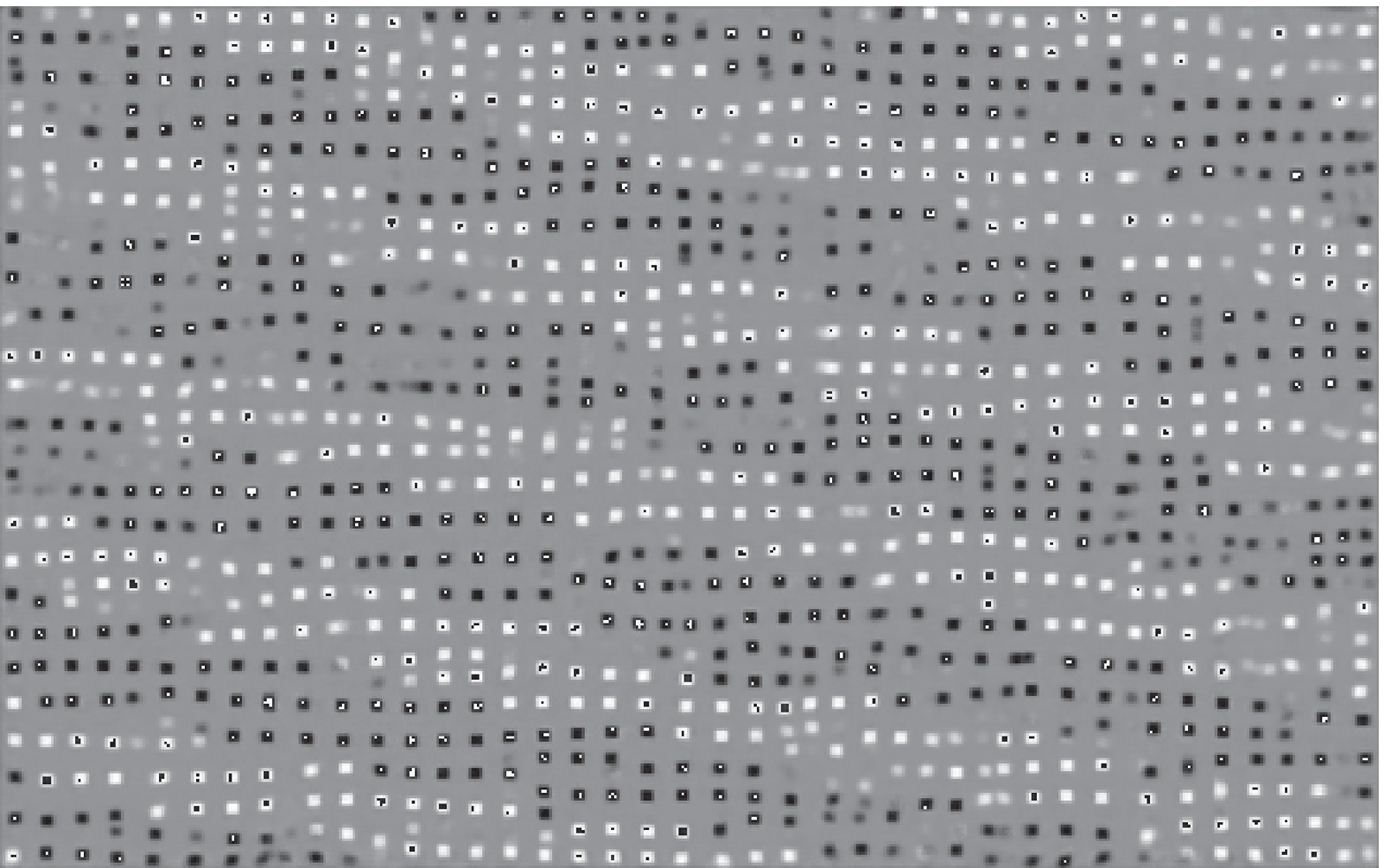}
\includegraphics[width=0.3\linewidth]{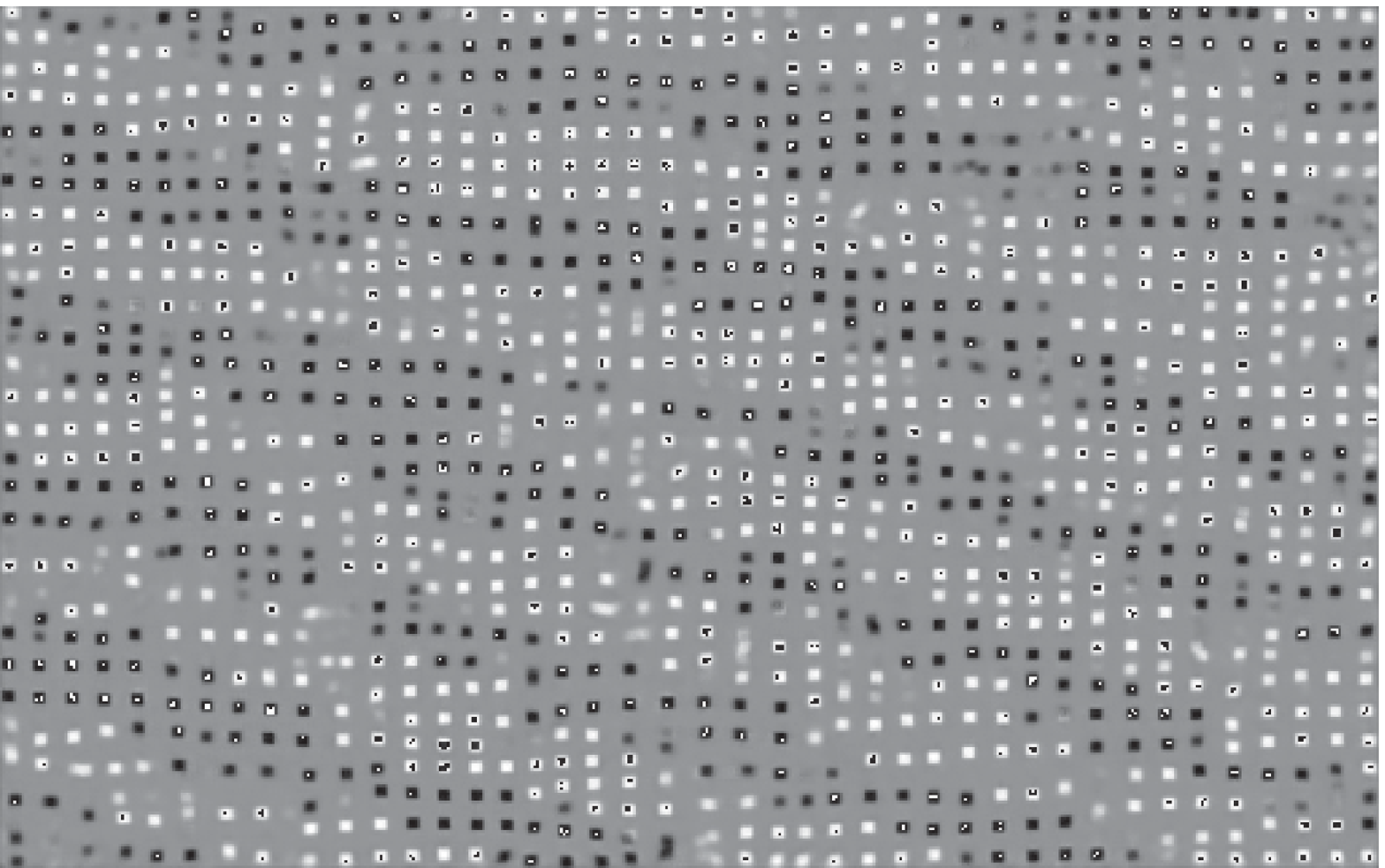}
}
\quad
\subfigure[\id{(\#75)}\add{ROC-like curve.}]{
\includegraphics[width=0.3\linewidth]{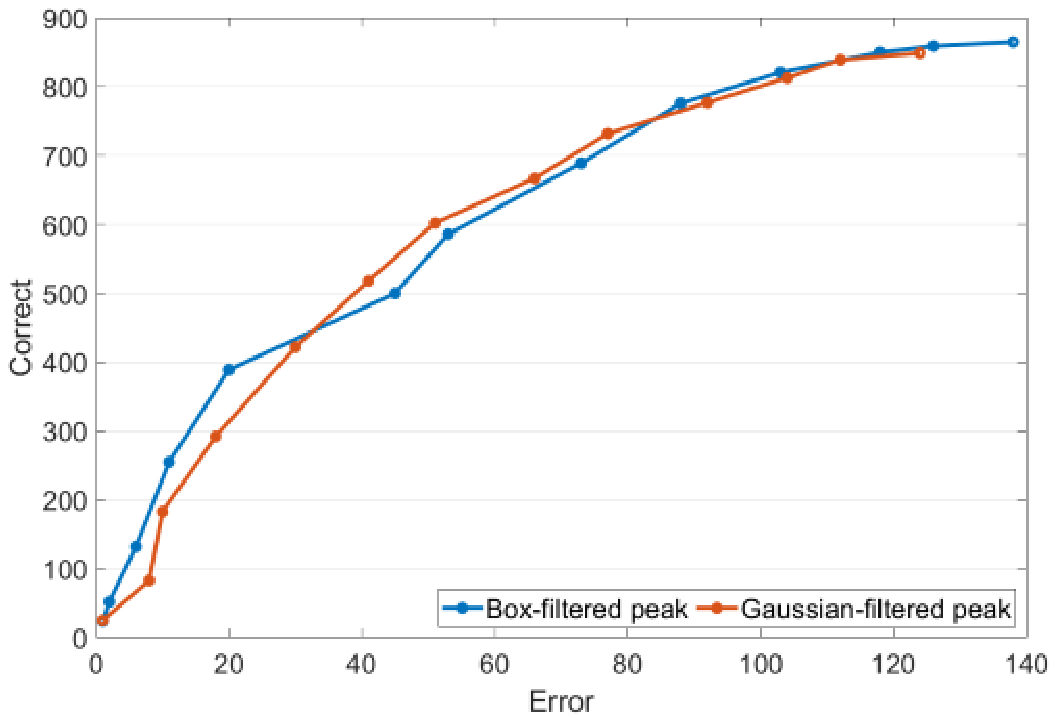}
\includegraphics[width=0.3\linewidth]{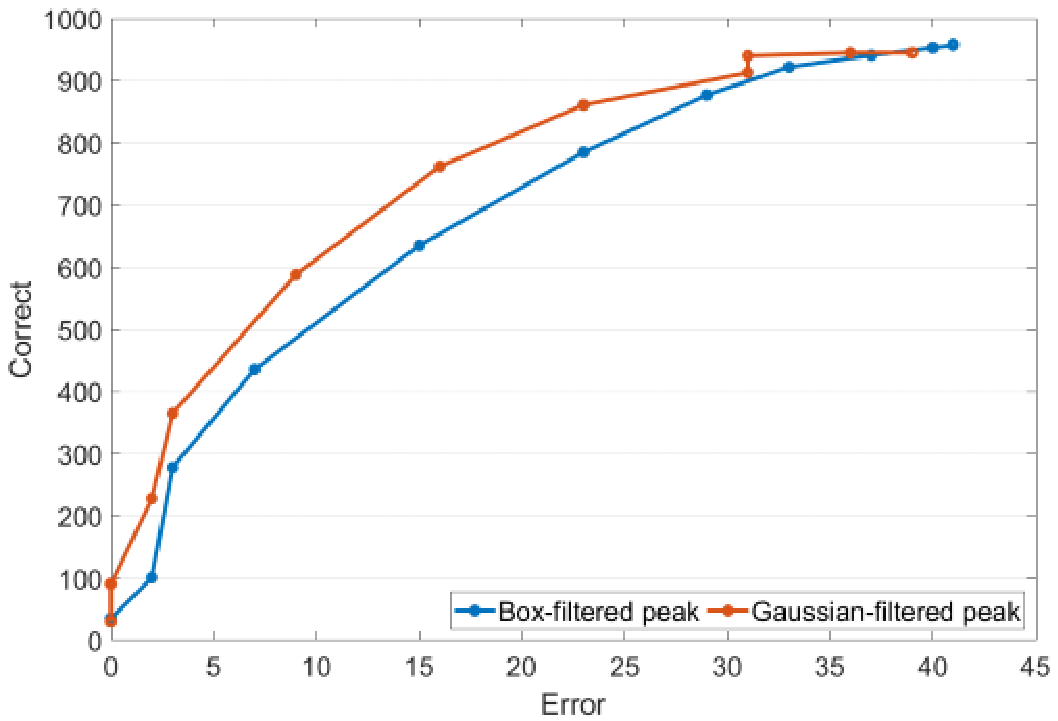}
\includegraphics[width=0.3\linewidth]{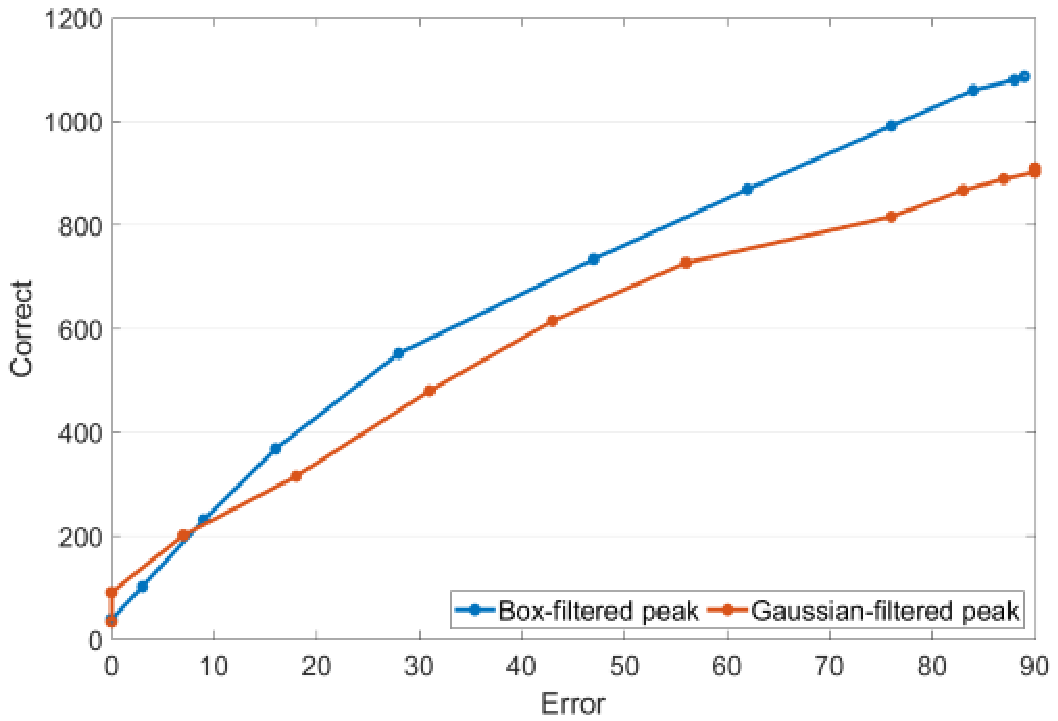}
}
\caption{Comparison of the different kinds of intermediate representations for DNN training. \id{\#64}\add{The results shown in (c) are given with $\sigma=5$ for Gaussian filter, and (d) are given with $N(x,y)$ as a window of $9 \times 9$ pixels. } }
\label{fig:comparison}
\end{figure*}

Textile samples were observed with an SLR camera Canon EOS M for consumer use; we  attached a fixed focus macro lens with LED illumination Canon EF-S 35mm f/2.8 Macro IS STM. High-resolution images were divided into small, partial images. 
\id{(\#32)}\add{The images were made by observing samples represented in reference \cite{Toyoura19}. The samples were woven with black and white yarns only, and the patterns were generated from natural images. }
\id{(\#18, \#57, \#58, \#67, \#72)}\add{We obtained 176 images; each had a resolution of $512 \times 320$ pixels. We then manually tagged all the images.}
\delete{We obtained $224$ images of $512 \times 320$ pixels using $176$ images for training and $48$ for validation.} \id{(\#54)}\add{The number of images is to be increased to get more accurate results, but it requires more effort. }
The training images were augmented by flipping them vertically and horizontally and rotating them $180^{\circ}$; then, $704$ \id{(\#22, \#32, \#67)} \delete{augmented images were used}\add{examples were used for the training. With the horizontal mirror images, the vertical mirror images, the images flipped $180^{\circ}$, and the original images, there were a total of $176 \times 4 = 704$ images.}
For each image, we performed the manual cross point extraction as described in Section 3.1. \id{(\#22)}\add{The observed images and manually added tags are shared at  https://github.com/toyoura/TextileDecoding.} 

PyTorch was employed as a deep learning framework and GPU GeForce2080Ti 11GB was used for training and validation. It took about $2.4$ hours to train. The batch size was set to $4$ and the maximum epoch was set to $400$. To complete the process of one image at runtime it took about $1.21$ seconds, which included $0.24$ second for pre-processing, $0.08$ second for executing DNN, and $0.89$ second for post-processing.

\subsection{Comparison of intermediate representations}

DNNs were respectively trained by the three intermediate representational images described in Section 3.2. 
\id{(\#18, \#57, \#58, \#67, \#72)}\add{We performed an 11-fold cross-validation procedure to ensure the robustness of the method. The 176 images were divided into 11 groups of 16 images. After training the network with 160 images, we verified the accuracy of the remaining 16 images and repeated this process 11 times.} The output images obtained by each training are shown in Fig. \ref{fig:comparison}. For the results, no post-processing was done.

First, we can see that impulse peak pattern training does not provide good results. Many regions of the white yarns were detected as crossing points, and candidate points also appeared at the edges formed by the yarn contours. When considering the segmentation of post-processing, the noisy clutters did not result in the correct detection of candidate points.

Although much better results were provided than those by impulse peak pattern training, Gaussian-filtered peak pattern training generated blurry images. The post-processing of the segmentation was not successful because each of the regions became one continuous region of adjacent, segmented regions. This trend did not improve, even after changing the variance of the Gaussian peak. When the variance was set too small, the results were similar to those induced by the impulse peak pattern.

Finally, the results trained by the box-filtered peak pattern were better than the two previous results. The region indicating each crossing point was less likely to be integrated with its adjacent regions. The extraction of the candidate representative points by post-processing also showed good results in appearance.
\add{The correct rate for all groups was $83.25 \%$ on average, and the minimum was $74.13 \%$, the maximum was $99.17 \%$, and the standard deviation was $8.40 \%$. }

\subsection{Performance of cross point detection}

The resulting binary patterns obtained in the form of matrices could not be evaluated directly because the same number of crossing points in each row and column could not be aligned in the observed image. Therefore, the observed crossing points could not be represented in a matrix manner. Instead of the direct evaluation of the binary patterns, we verified whether the label images were correctly obtained. The correct label image could be given by manual labeling. The accuracy of the position of the crossing points could also be examined.

\begin{figure}[ht]
\centering
\includegraphics[width=0.47\textwidth]{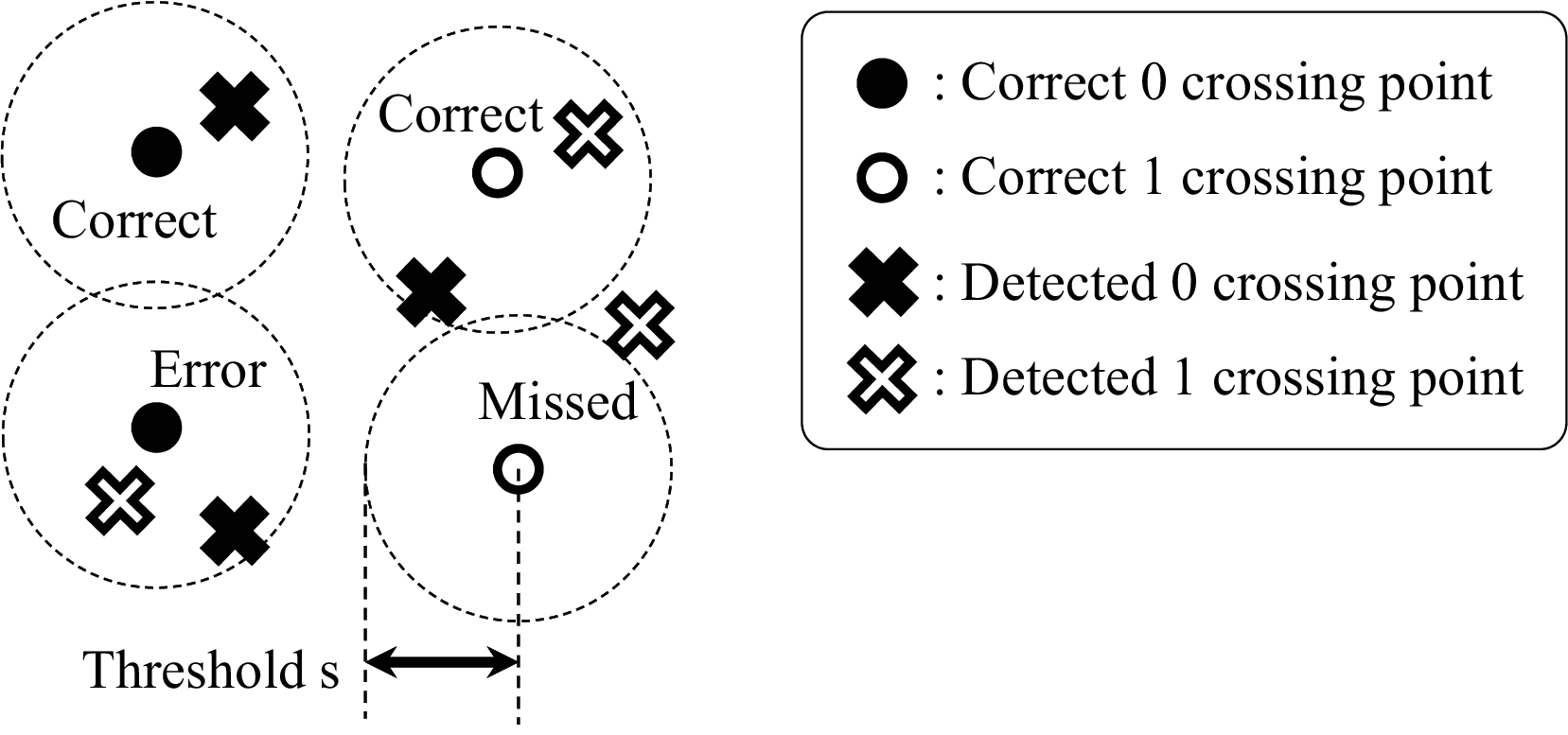}
\caption{Threshold $s$ for quantitative performance assessment.}
\label{fig:th_s}
\end{figure}

Here, we evaluated the representative points of the crossing point regions obtained from the resulting label image. Many pixels took a value of $0.5$ and a few took $0$ or $1$ to indicate that the pixels were representative crossing points. Similarly, in the manually labeled images, a small number of crossing points were placed in a background of many pixels that took a value of $0.5$. In order to verify the match between them, we set a threshold $s$ for Euclidean distance. For a crossing point in the manually tagged image, if the extracted candidate crossing points were within the threshold distance, we considered the crossing point to be correctly detected; if there were multiple intersections within $s$, we considered the crossing point with the smallest distance to be the corresponding point as shown in Fig. \ref{fig:th_s}. 

\begin{figure*}[ht]
\centering
\subfigure[Number of layer.]{
\begin{minipage}[t]{0.18\linewidth}
\centering
\includegraphics[width=1in]{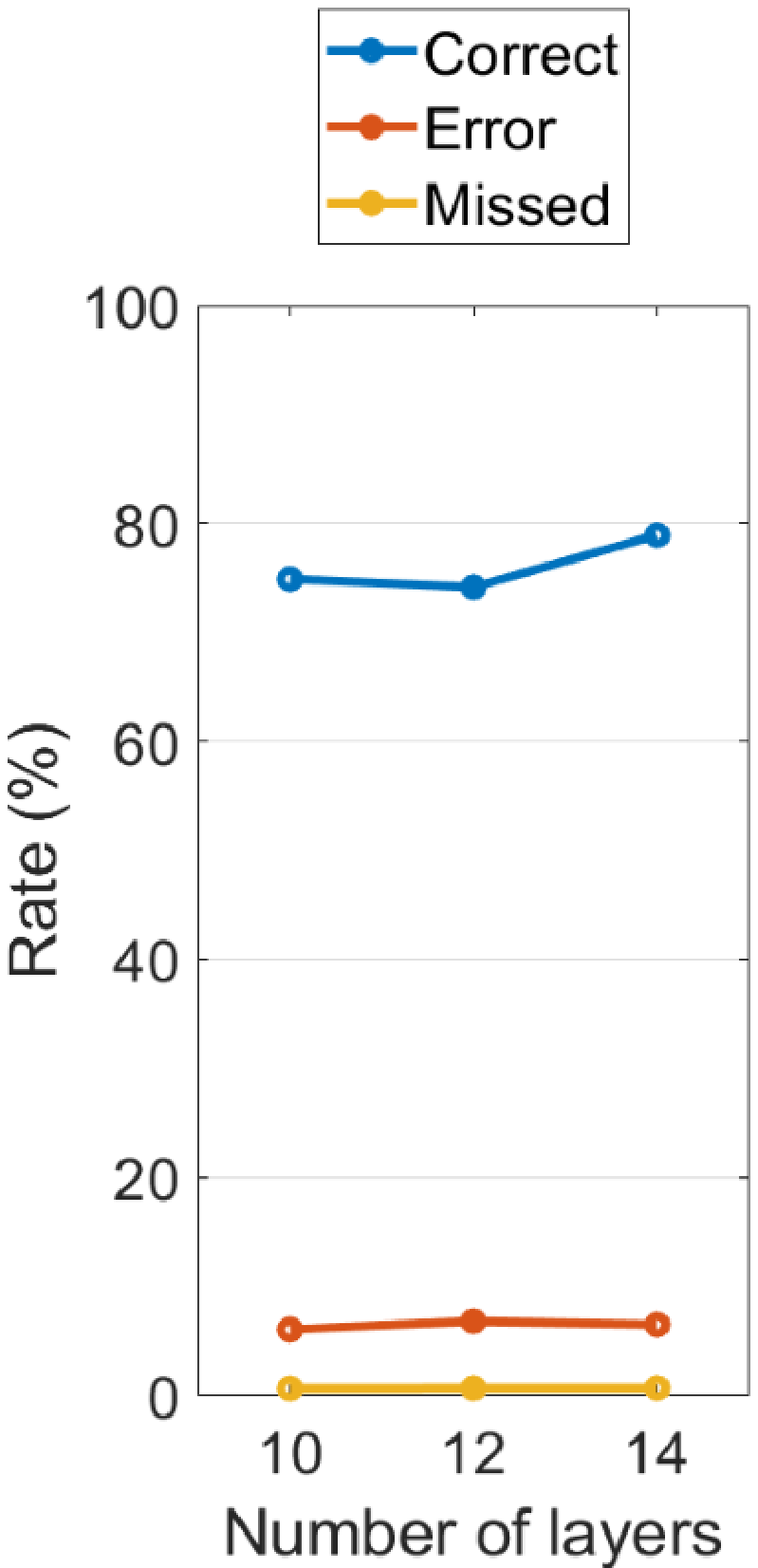}
\end{minipage}
}
\subfigure[Box filter size.]{
\begin{minipage}[t]{0.18\linewidth}
\centering
\includegraphics[width=1in]{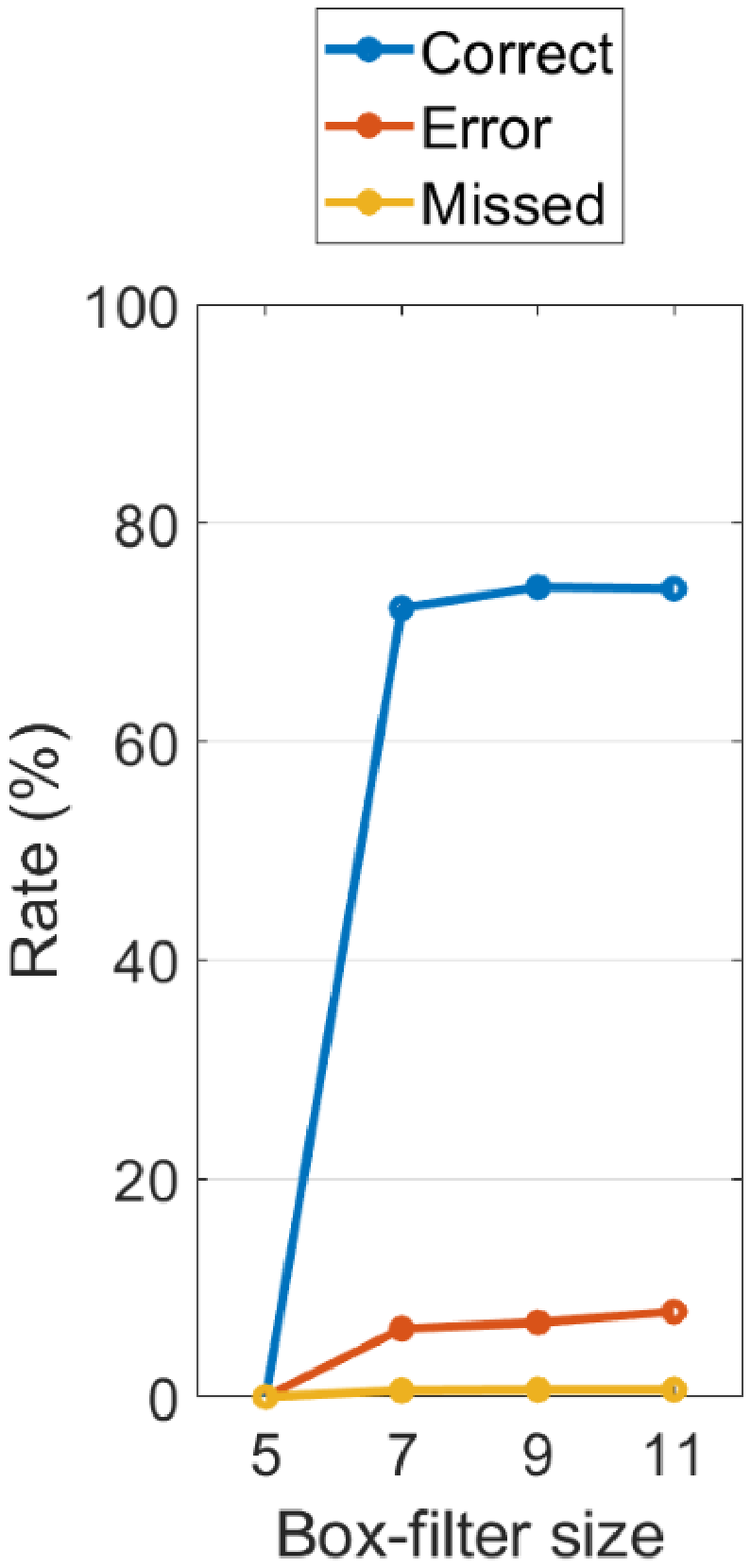}
\end{minipage}
}
\subfigure[Gaussian filter size.]{
\begin{minipage}[t]{0.18\linewidth}
\centering
\includegraphics[width=1in]{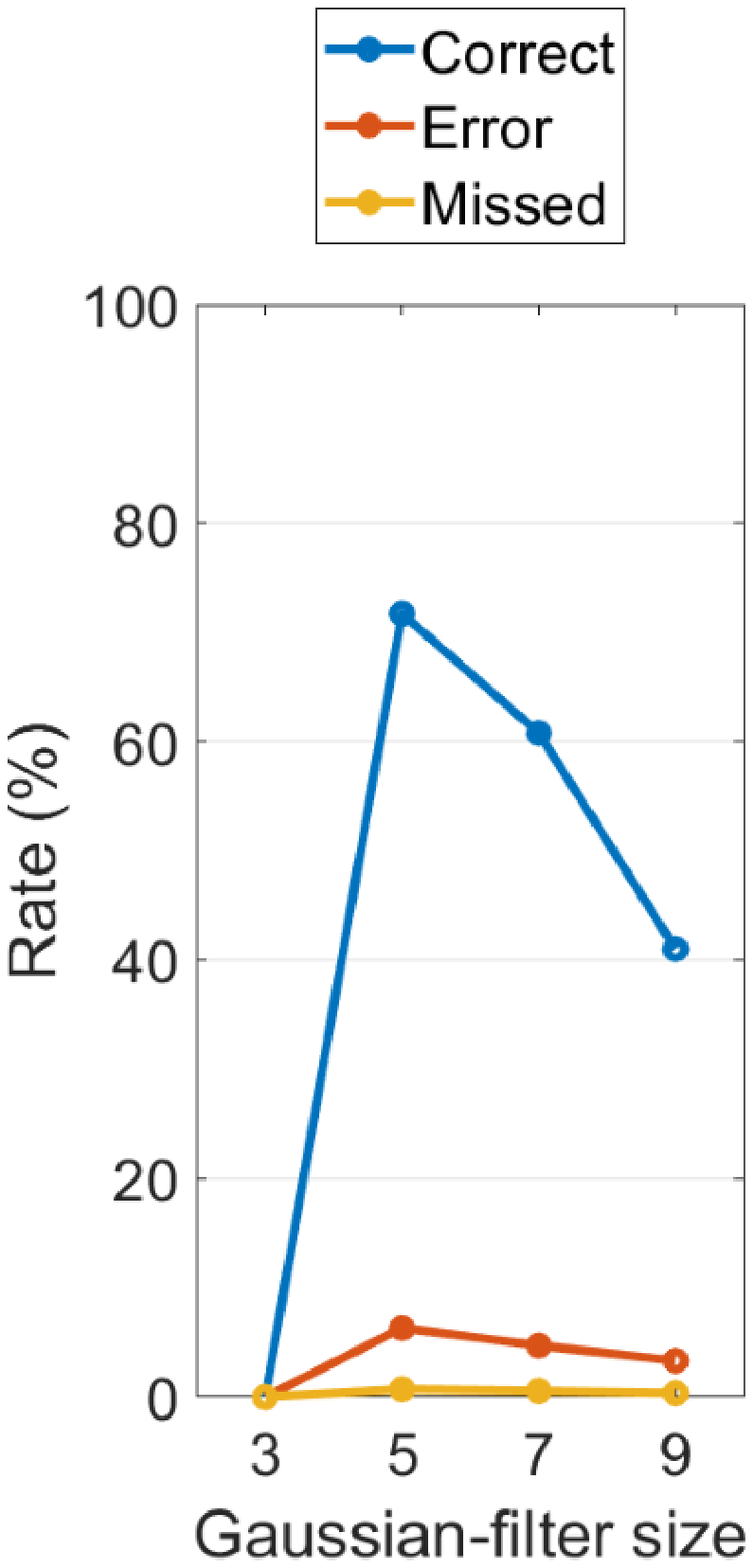}
\end{minipage}
}
\subfigure[s in Box filter.]{
\begin{minipage}[t]{0.18\linewidth}
\centering
\includegraphics[width=1in]{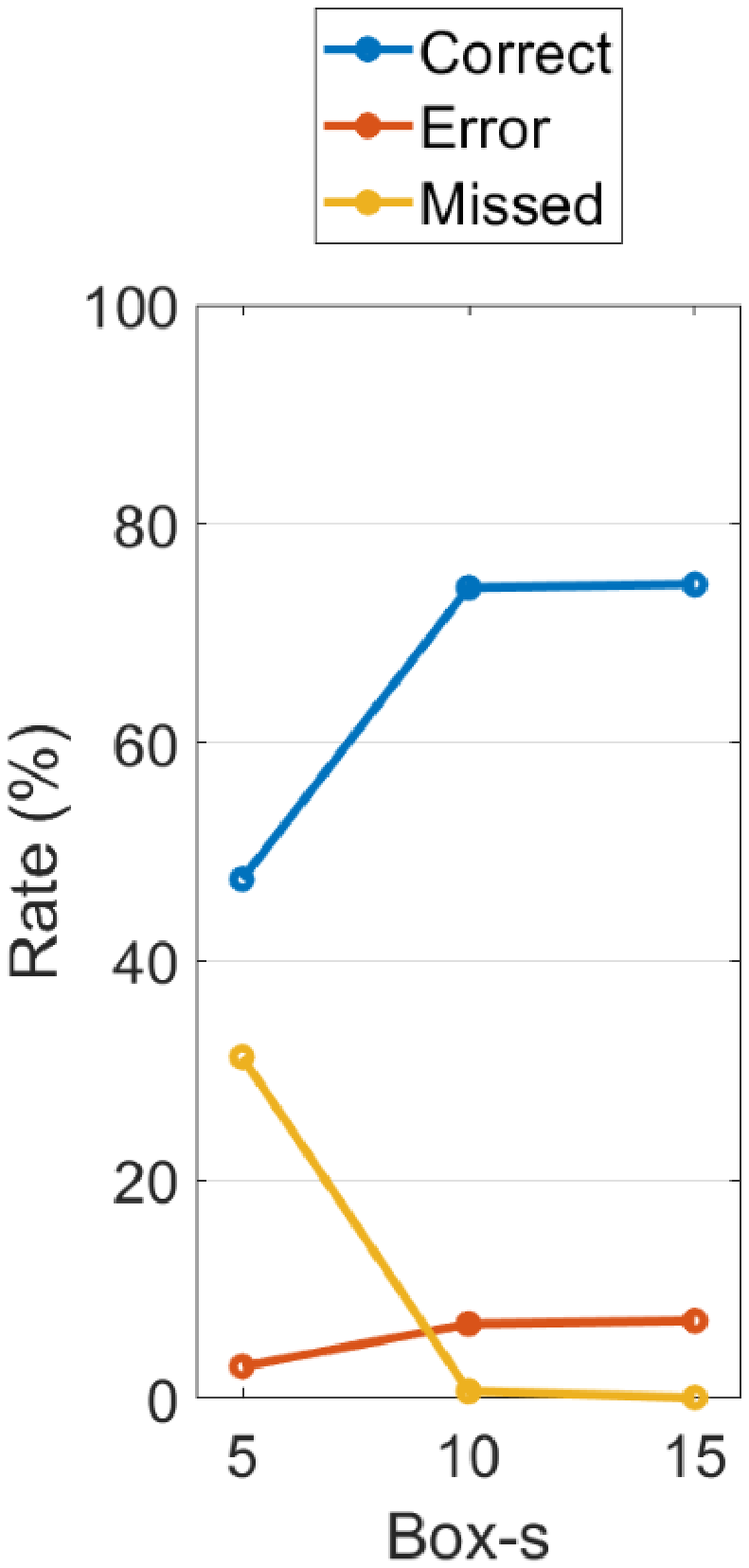}
\end{minipage}
}
\subfigure[s in Gaussian filter.]{
\begin{minipage}[t]{0.18\linewidth}
\centering
\includegraphics[width=1in]{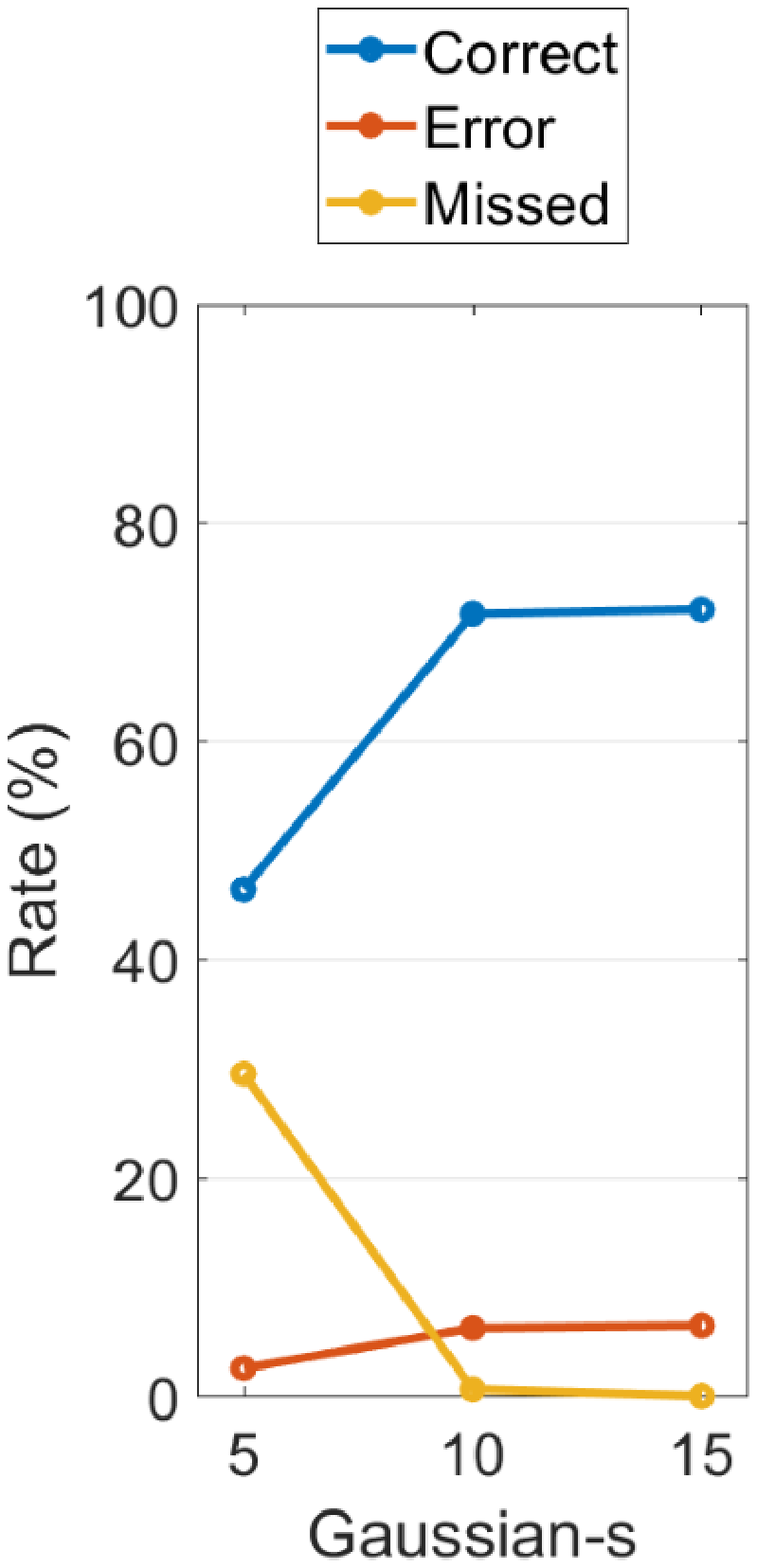}
\end{minipage}
}
\centering
\caption{\add{ Comparison experiment for different number of layers, filters, and distance threshold s.}}
\label{fig:analysis}
\end{figure*}

\begin{figure*}[ht]
\centering
\subfigure[Input image.]{
\begin{minipage}[t]{0.22\linewidth}
\centering
\includegraphics[width=1\linewidth]{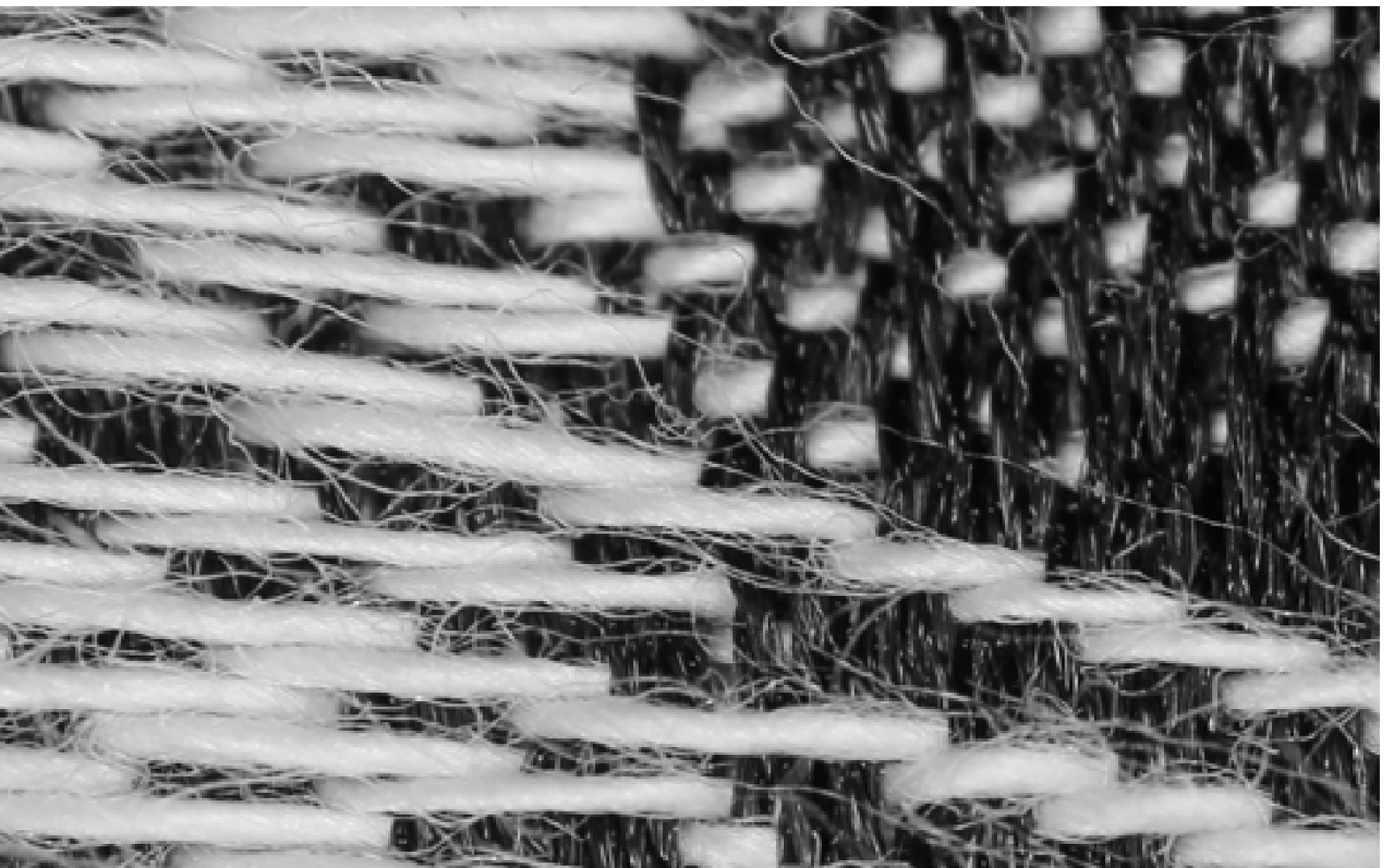}\vspace{4pt}
\includegraphics[width=1\linewidth]{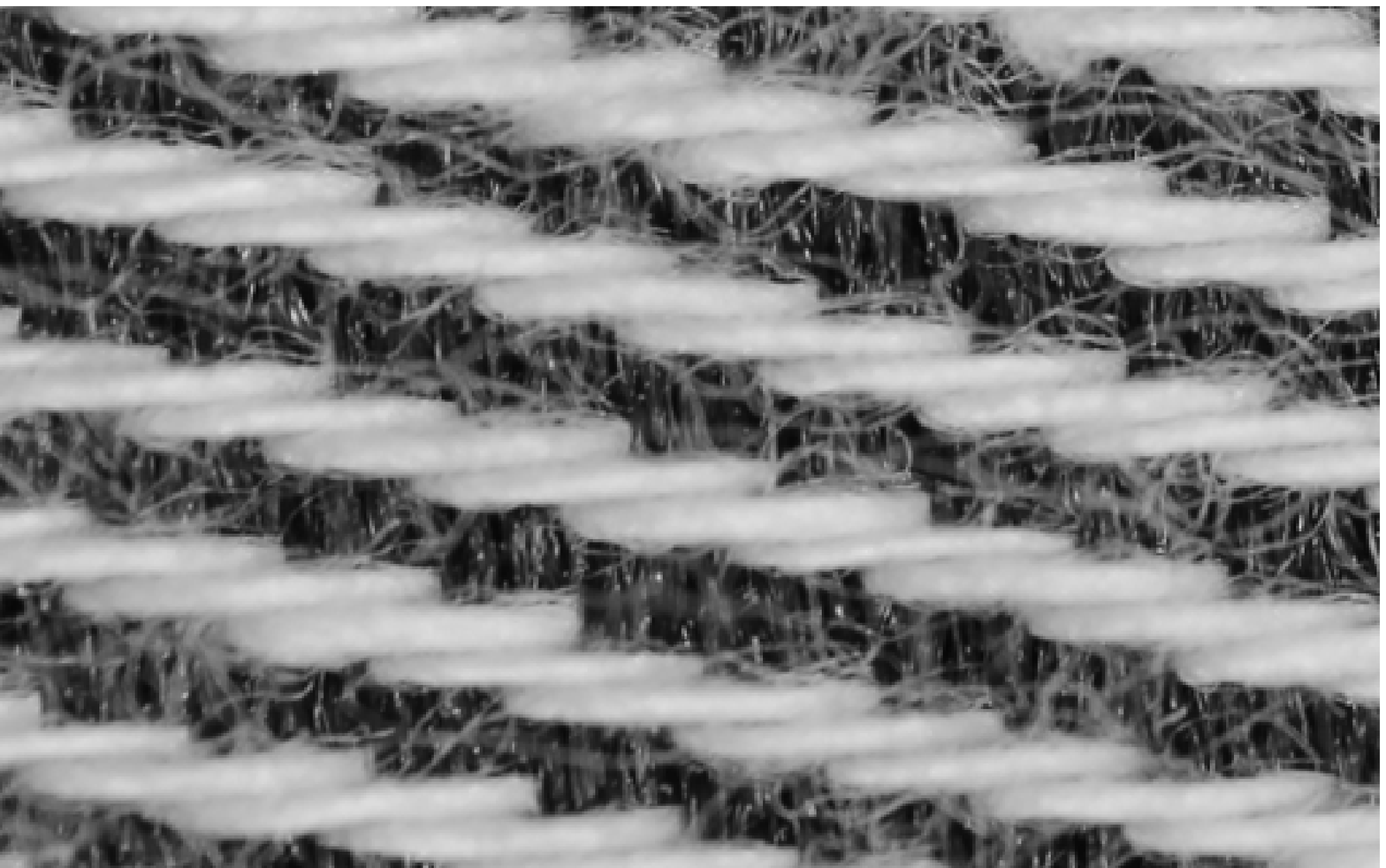}\vspace{4pt}
\includegraphics[width=1\linewidth]{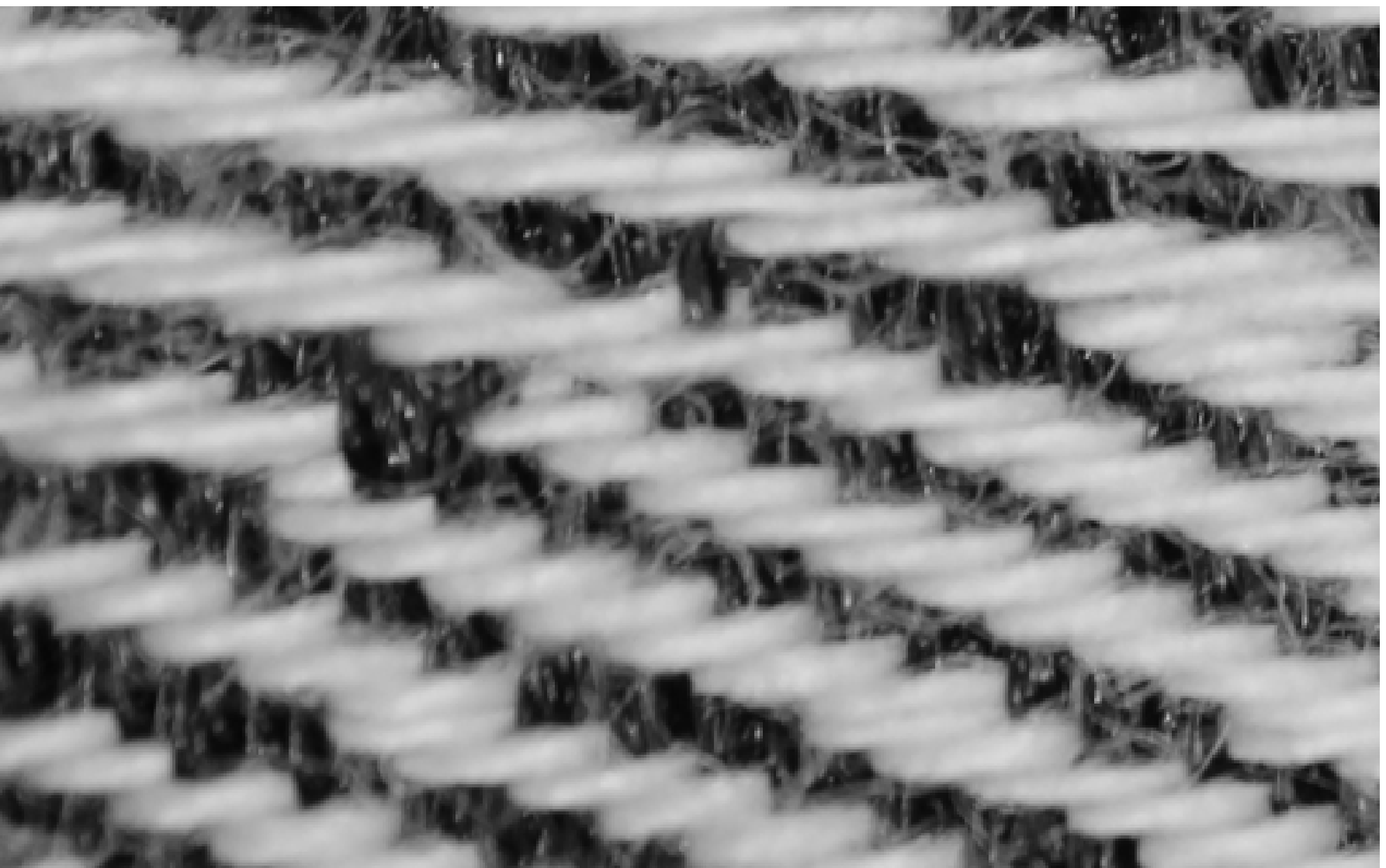}\vspace{4pt}
\includegraphics[width=1\linewidth]{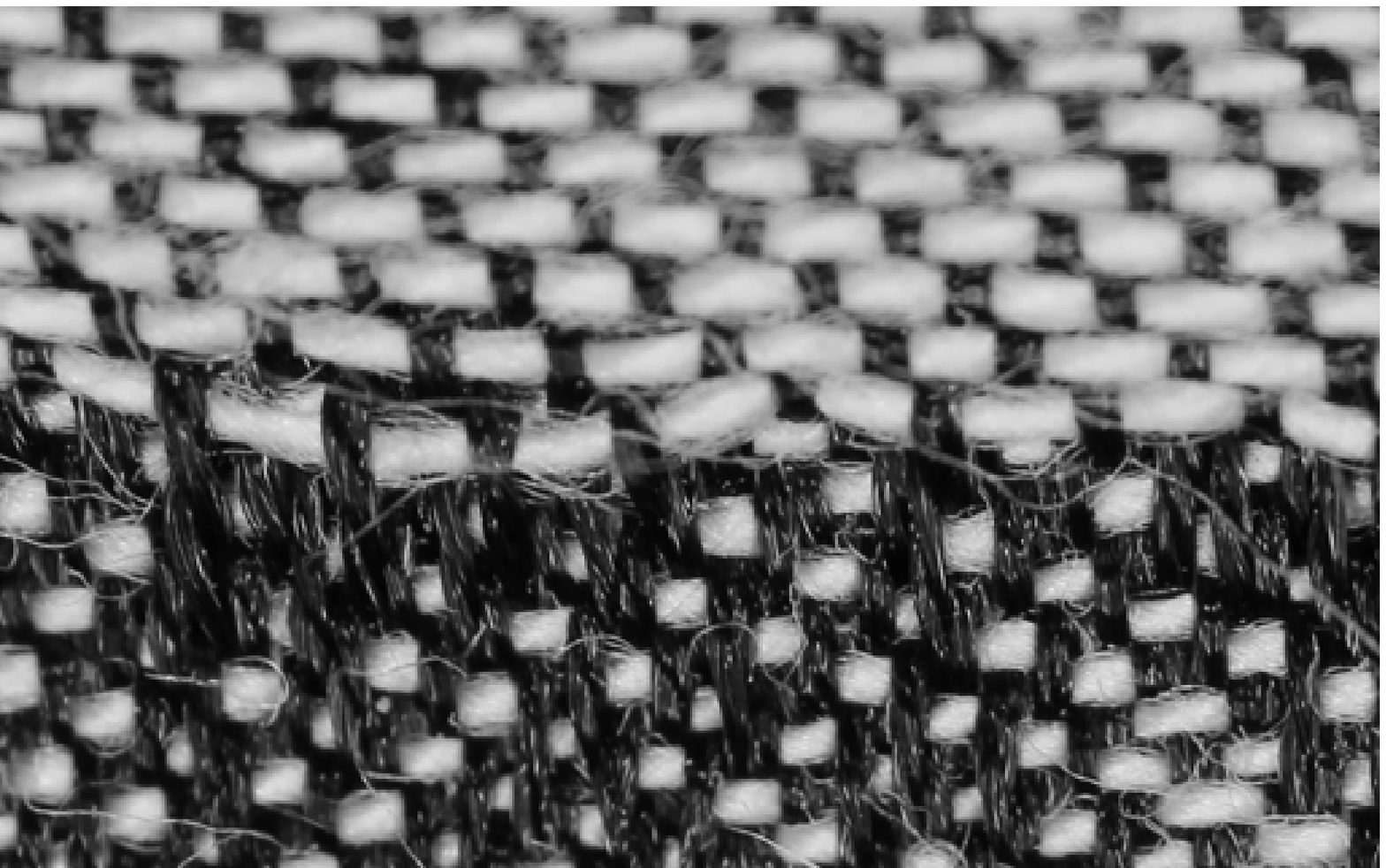}\vspace{4pt}
\end{minipage}}
\subfigure[Estimated binary pattern.]{
\begin{minipage}[t]{0.22\linewidth}
\centering
\includegraphics[width=1\linewidth,height=0.625\linewidth]{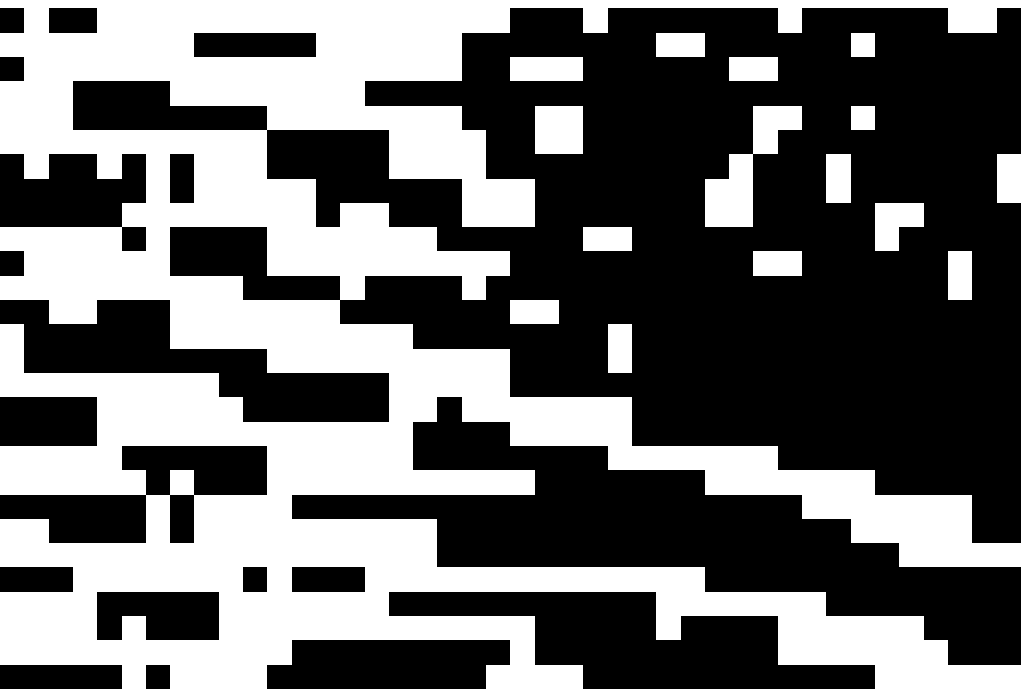}\vspace{4pt}
\includegraphics[width=1\linewidth,height=0.625\linewidth]{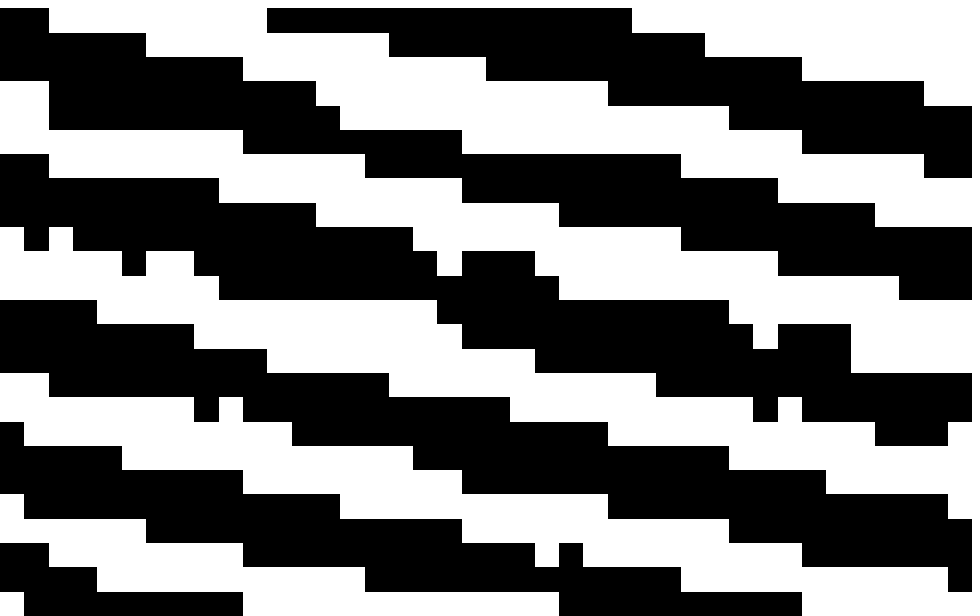}\vspace{4pt}
\includegraphics[width=1\linewidth,height=0.625\linewidth]{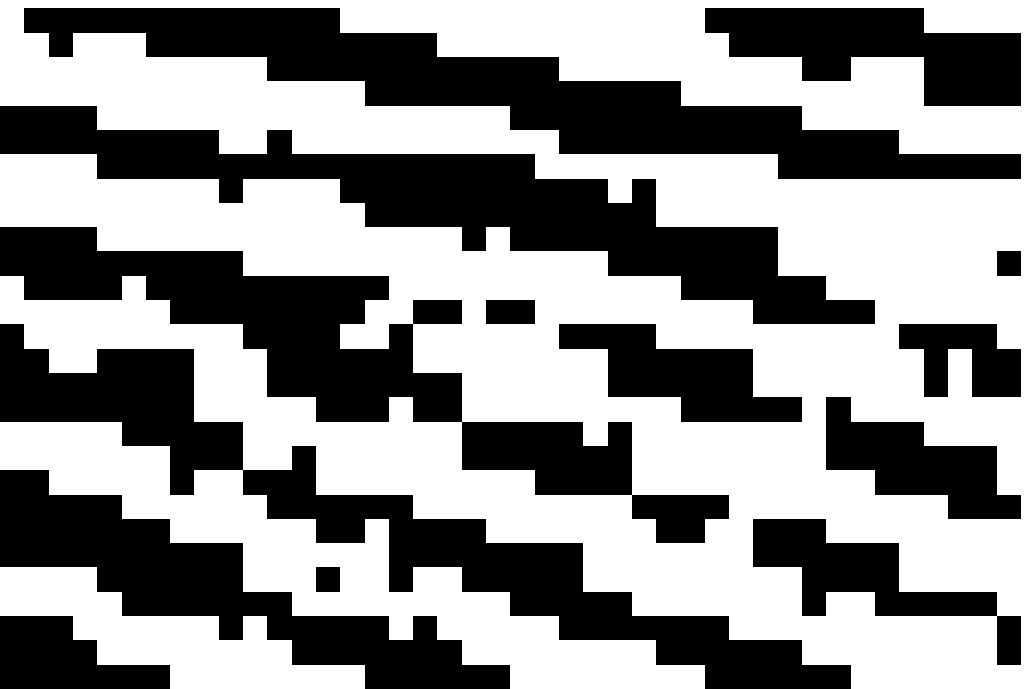}\vspace{4pt}
\includegraphics[width=1\linewidth,height=0.625\linewidth]{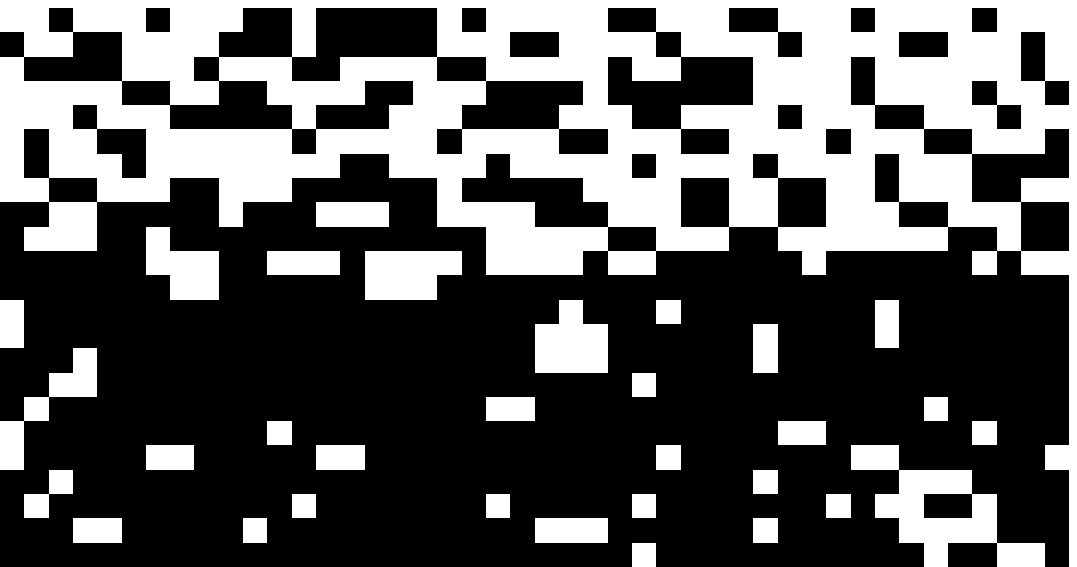}\vspace{4pt}
\end{minipage}}
\subfigure[\add{Binary pattern validation.}]{
\begin{minipage}[t]{0.22\linewidth}
\centering
\includegraphics[width=1\linewidth,height=0.625\linewidth]{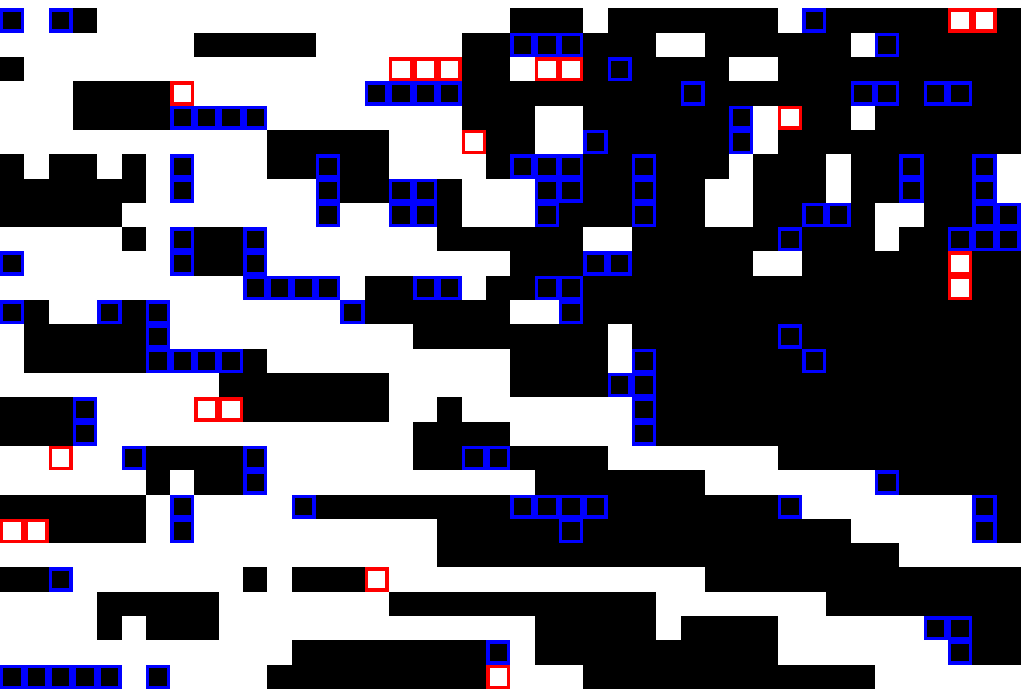}\vspace{4pt}
\includegraphics[width=1\linewidth,height=0.625\linewidth]{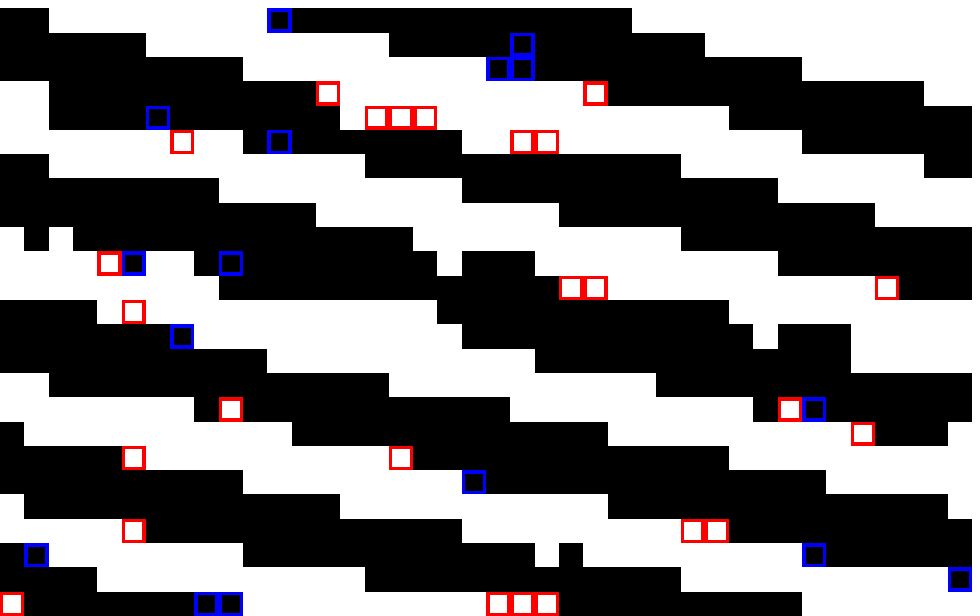}\vspace{4pt}
\includegraphics[width=1\linewidth,height=0.625\linewidth]{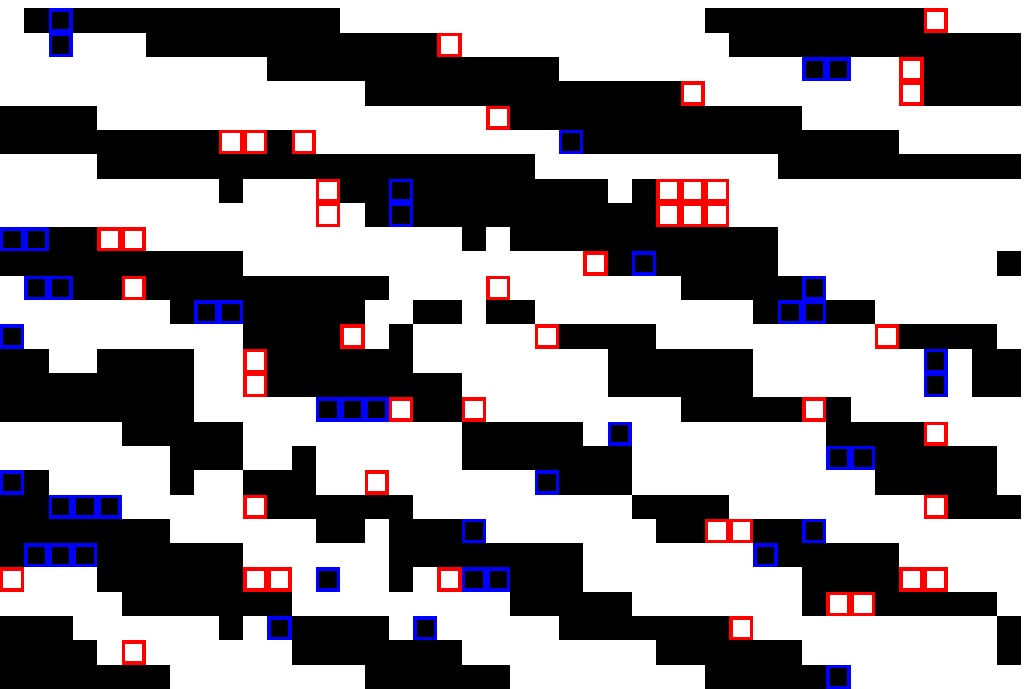}\vspace{4pt}
\includegraphics[width=1\linewidth,height=0.625\linewidth]{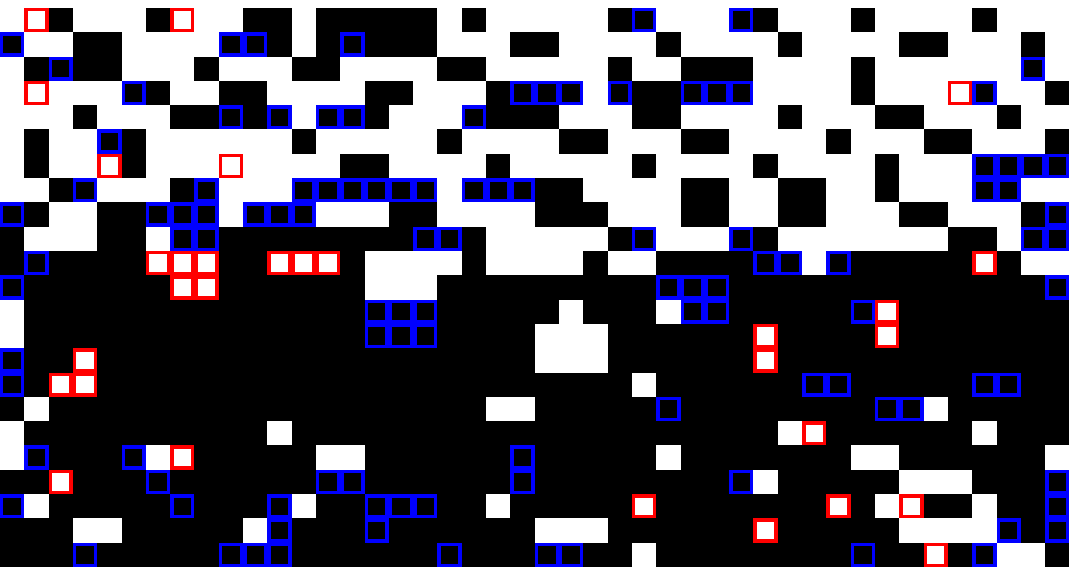}\vspace{4pt}
\end{minipage}}
\subfigure[Reproduced woven fabric of (b).]{
\begin{minipage}[t]{0.22\linewidth}
\centering
\includegraphics[width=1\linewidth]{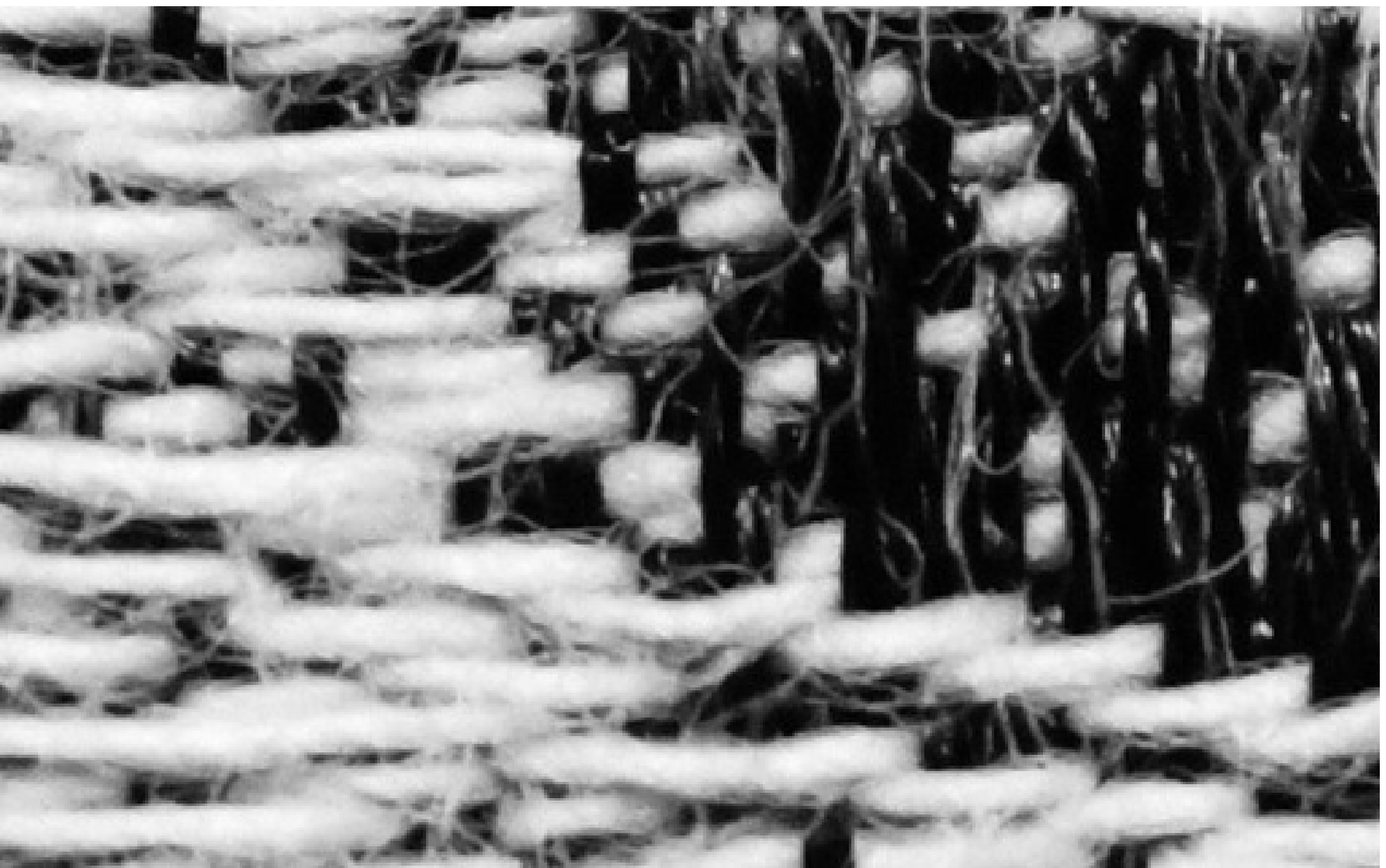}\vspace{4pt}
\includegraphics[width=1\linewidth]{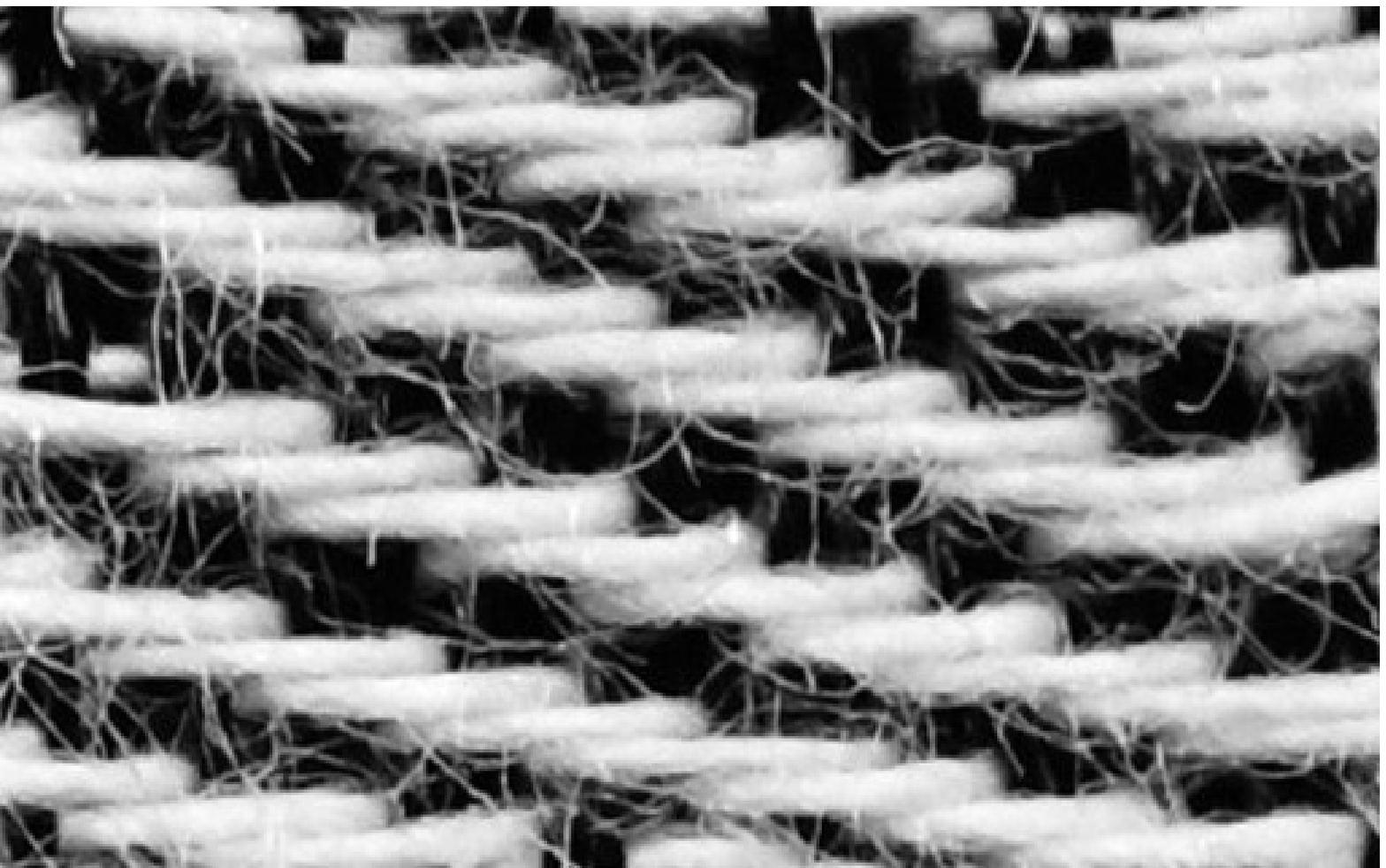}\vspace{4pt}
\includegraphics[width=1\linewidth]{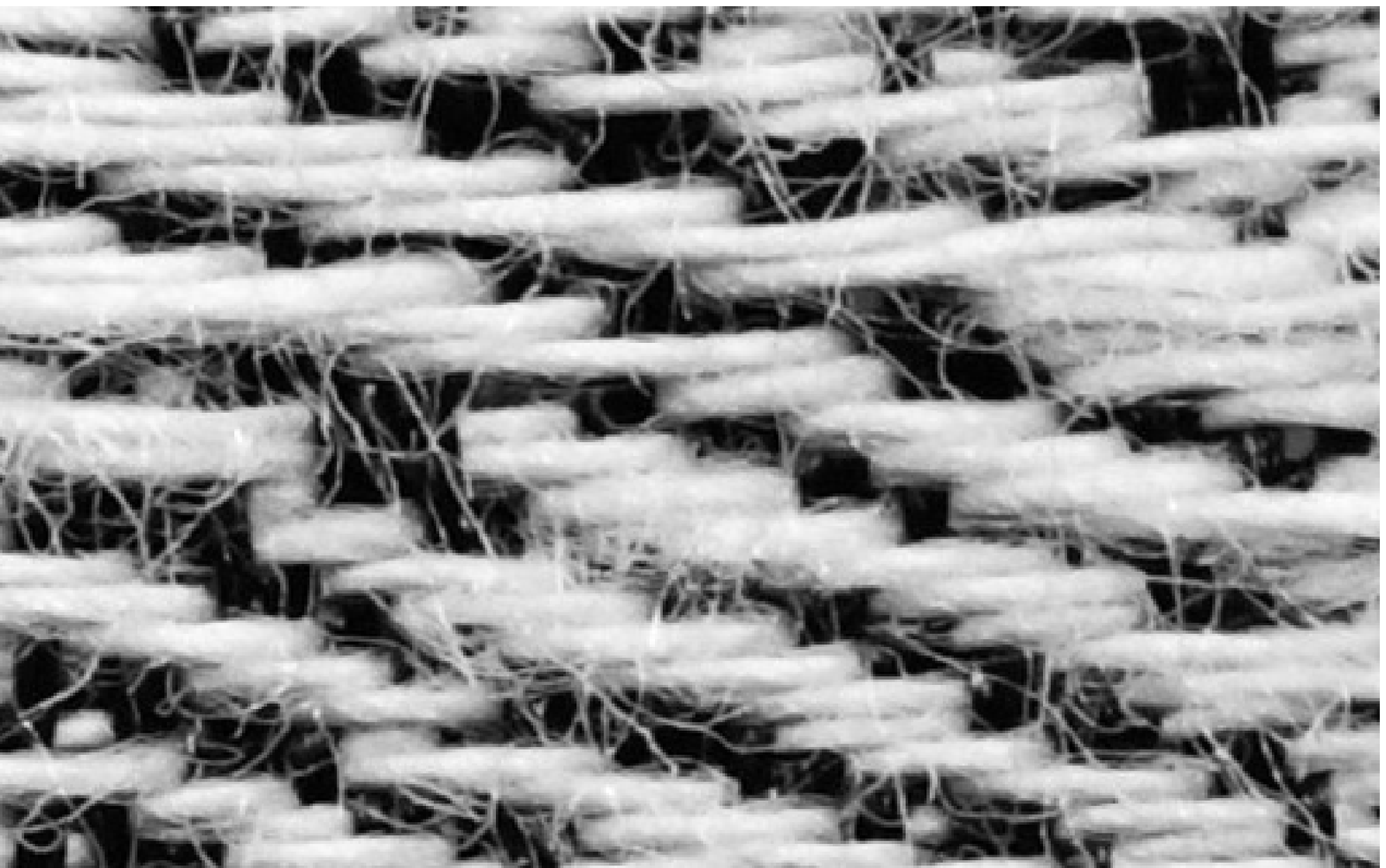}\vspace{4pt}
\includegraphics[width=1\linewidth]{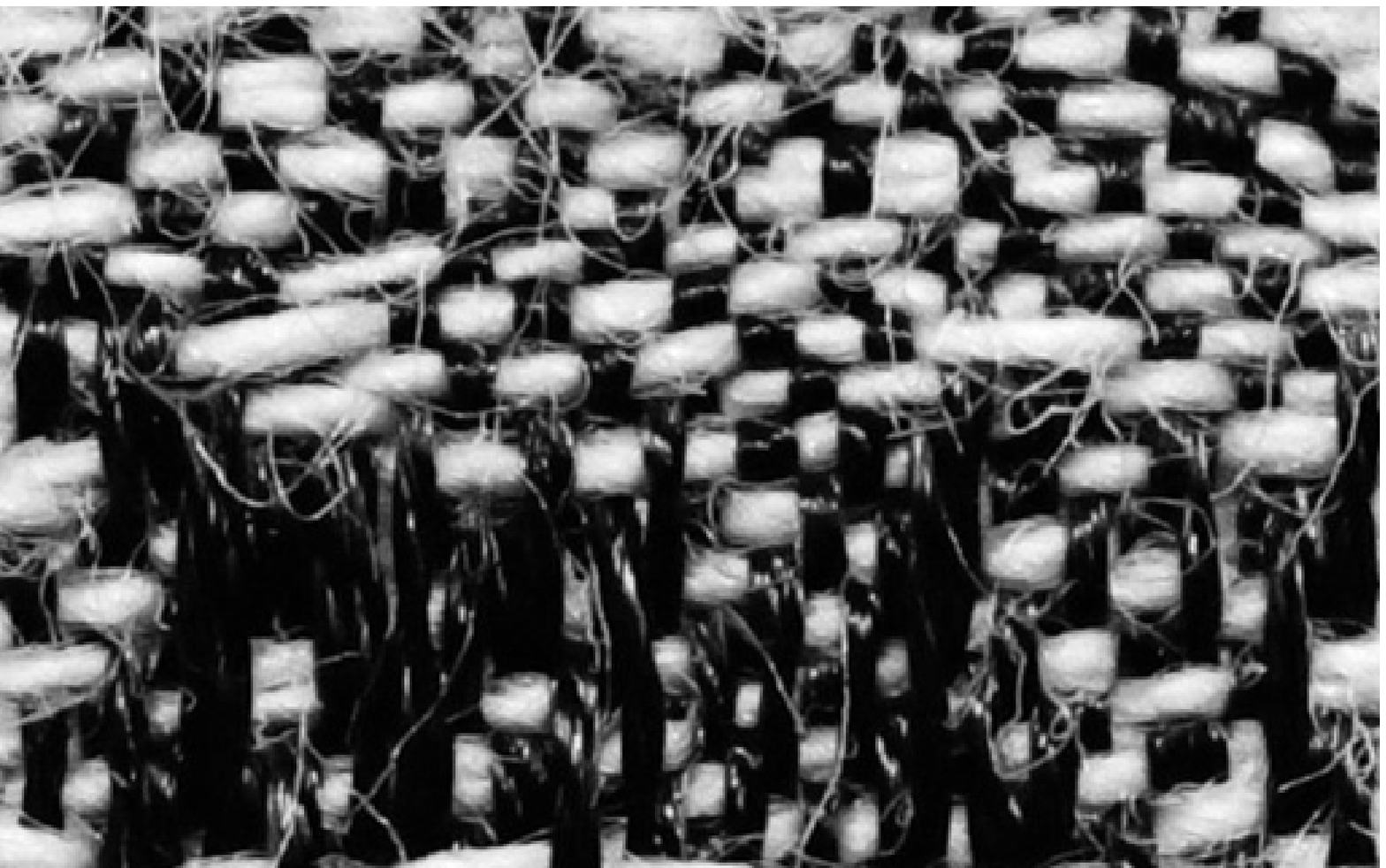}\vspace{4pt}
\end{minipage}}
\centering
\caption{Reproduced woven fabric by final decoded binary patterns.}
\label{fig:result02}
\end{figure*}

For quantitative evaluation, for each representative crossing point in the \add{estimated binary pattern}\delete{manually labeled image}, the rate of correctly detected crossing points (Correct), the rate of incorrectly detected crossing points (Error), and the rate of points with no candidate points in distance $s$ (Missed) were calculated. These rates varied depending on the value of $s$ and were represented by ROC-like curves.

Fig. \ref{fig:comparison}(e) shows the resulting label images and ROC-like curves. Note that we omitted the quantitative analysis for impulse peak pattern training. There were not enough extracted crossing points from impulse peak pattern training because resulting output images had few large white and black regions after post-processing.

\id{(\#12,\#19,\#26,\#29,\#30,\#71,\#76)}\add{After extensive trial and error, we found a DNN with 12 layers, a $9 \times 9$ box-filter, and a distance threshold $s = 10$ indicated good performance. In the following, we show the experimental results for different filters, the window size, the number of layers, and the distance threshold s. From the image set used for the 11-fold cross-validation, we employed the worst group (160 images for training and 16 images for test) for the comparison experiment. }

\vspace{0.2cm} 
\noindent {\it \id{(\#65,\#79)}{\add{Number of DNN layers}}}
\add{To confirm the optimal structure of the DNN, we designed comparative experiments,  reducing and increasing the number of neural network layers. The number of layers should be even because the same number of layers is required for encoder and decoder parts. The deepest layers near the bottleneck were added and removed for changing the number of layers. The smallest layer of 14-layer network is $5 \times 8 \times 48$, then we cannot add any more layers to 14-layer network. Fig. \ref{fig:analysis}(a) shows that the 12-layer neural network had an accuracy level of $74.13\%$. Under the same circumstances, the 10-layer neural network had an accuracy level of $74.88\%$, while the accuracy of the 14-layer neural network was $78.92\%$. For the size of the dataset and the degree of difficulty of the task, the 14-layer network had the best performance. }

\vspace{0.2cm} 
\noindent {\it {\add{Filter parameters}}}

\add{For the intermediate representation, we controlled other unchanged variables, and compared their results by changing the size of the filter. Although there were a number of values from which to choose, considering the feasibility of the experiment, as shown in Figs. \ref{fig:analysis}(b) and \ref{fig:analysis}(c), we compared the Box methods with filter sizes of 5, 7, 9, and 11; we compared the Gaussian methods with filter sizes of 3, 5, 7, and 9. When the filter size was 9, the Box-filter method had the highest accuracy of $78.92\%$, 7 and 11 had accuracy levels of $72.21\%$ and $73.99\%$, respectively. When the filter size was 5, the Gaussian-filter method had the highest accuracy of about $71.69\%$; 7 and 9 had accuracy levels of about $60.77\%$ and $41.00\%$, respectively. For Gaussian size 3 and Box size 5, we could not extract the binary pattern, so their accuracy was 0. By comparison, when the filter size of the Box-filter method was 9, that yielded the best result, and that was the scheme adopted for our method.}

\vspace{0.2cm} 
\noindent {\it {\add{Distance threshold $s$}}}

\delete{Each resulting label image showed the same trend, with approximately $90 \%$ of correct crossing points at $s=8$; $95 \%$ of correct crossing points at $s=13$; less than $0.1 \%$ of error crossing points at $s=13$; and only $5 \%$ of missed crossing points when $s$ was increased. This indicated that $95 \%$ of the crossing points were detected correctly within the distance $s$, while $5 \%$ of the warp and weft crossing points were detected incorrectly. Statistically, as shown in Table 1, the rate of correct crossing points was greatest at $s=13$, which was $94.09 \%$.}
\add{Each resulting label image showed the same trend, as shown in Figs. \ref{fig:analysis}(d) and \ref{fig:analysis}(e), there were $47.47\%$ of correct crossing points at $s = 5$; $74.13\%$ of correct crossing points at $s = 10$; a $6.81\%$ error crossing points at $s = 10$; and only $0.67\%$ of missed crossing points when $s$ was increased. This indicated that $74.13\%$ of the crossing points were detected correctly within the distance $s$, while $7.48\%$ of the warp and weft crossing points were detected incorrectly. Statistically, the rate of correct crossing points was greatest at $s=15$, which was $74.44\%$. Considering the total rate of correct, error, and missed, the performance is the best at $s = 10$. }

\subsection{Validation of decoded binary patterns}

Fig. \ref{fig:result02} shows the results that were woven by the obtained binary pattern, together with the observed image. Since the input image contained crossing points near the image boundary that could not be expressed in the form of a matrix, we ignored the intersections near the boundary and evaluated them. \id{(\#20, \#34, \#40, \#59, \#72)}\add{In addition, we added binary pattern validation in Fig. \ref{fig:result02}(c) to show the analysis results. We evaluated whether the values of 0 and 1, given the grid points in the estimated binary image matched the values in the ground truth image. Each grid point corresponded to a pixel of the observed image, so we could check the match between it and the binary value of the nearest crossing point in the ground truth image. In Fig.  \ref{fig:result02}(c), a red box shows that 0 was wrongly estimated as 1, and a blue box shows that 1 was wrongly estimated as 0 for each grid point.} 
In the patterns, we were able to obtain results that were close in appearance. Since some of the patterns were extracted that are not common as woven patterns, such as too many or too few crossing, there was room for the pattern to be improved by converting it with an uncommon pattern such as a restriction. 
\id{(\#20, \#34, \#40, \#57, \#59, \#72)}\add{Although the error in which 0 was wrongly set to 1 and vice-versa is different from a false positive or a false negative in a general sense, the accuracy and F-measure of the statistical metric can be applied to the data. The accuracy for 176 images was 0.930 and the F-measure was 0.929, on average.}

Although the obtained pattern was not perfect, it can be said that the same pattern was approximately obtained automatically. 

\section{Conclusions}
In this paper, we proposed a method for decoding the binary patterns that define the weaving of fabric. By developing intermediate representations, we were able to accomplish the task by deep learning. The pre-processing and post-processing allowed us to bridge the intermediate representation image and the binary pattern. The experimental results showed that our method allowed for correctly extracting $93\%$ of the crossing points, and the reproduced textiles were close to the original one in appearance. 
\id{(\#50)}\add{Although the box filter may not be optimal, it gives better results than the isotropic Gaussian filter. We consider this to be due to the grid arrangement of the yarns.}

Only black and white yarn images were discussed in this paper. If the warp and weft are of one color each, the same processing is considered possible by converting the image according to the proximity of each color. Multiple colors may be used for weft yarns, and this will be handled by converting the image by the color distance from that of warp yarn; however we have not confirmed whether this will work well. That is a topic for future studies.

\idr{(\#1)}
It is necessary to ensure that each \add{of the pieces of} yarn \add{are}\delete{is} captured one by one in the observed image, which limits the scope of the observation. We need a method to integrate the resulting patterns observed at multiple locations. The resulting patterns are partially incomplete, which makes the problem more difficult. Various image stitching techniques are helpful. \id{(\#47)}\add{Although we used} 
\addr{LOG-filtered} 
\add{images for detecting the positions of the yarns, there is the possibility of improving the accuracy by introducing the structural tensor. In the case of textiles, the yarns are made of even finer yarns twisted together, so the fine edges can interfere with calculating the correct tensor.} We would like to address \delete{this issue} \add{these issues} in the future.

\id{(\#12, \#21, \#30, \#76)}\add{We will also leave the further tasks of the optimization of network configuration and the selection of the backbone network to future work. }
\idr{(\#3)}\addr{Recent powerful machine learning techniques \cite{Ahmadlou10,Rafiei17,Pereira20,Alam20} will contribute to improve the accuracy of pattern decoding even when the dataset is not large enough. }

\section*{Acknowledgement}
The authors appreciate the joint support for this project by the JSPS KAKENHI (Grant Nos. JP16H05867, JP20H04472), the NSFC-Zhejiang Joint Fund for the Integration of Industrialization and Informatization (Grant No. U1909210), the National Natural Science Foundation of China (Grant Nos. 61761136010, 61772163), \deleter{and} the Natural Science Foundation of Zhejiang Province (No.LY18F020016), \idr{(\#4)}\addr{and Zhejiang Lab Tianshu Open Source AI Platform (No.111007-AD1901)}.

\end{document}